%% file: main.tex
\documentclass[11pt]{style/prism-princeton}

\usepackage{titletoc}
\usepackage{graphicx}   
\usepackage{caption}    
\usepackage{listings}
\usepackage{wrapfig}

\newcommand{\method}{\textsc{RAVEN}\xspace}
\newcommand{\methodfull}{RetrievAl via Visual Embeddings for Navigation\xspace}
\newcommand{\vlmonly}{VLM Only\xspace}
\newcommand{\remembr}{ReMEmbR\xspace}
\newcommand{\navqa}{NaVQA\xspace}
\newcommand{\findingdory}{FindingDory\xspace}
\newcommand{\ourdataset}{\textsc{RAVEN-QA}\xspace}
\newcommand{\best}[1]{\textbf{#1}}

\input{preamble/notation}

\author[1*]{Yixun Hu}
\author[1*]{Zhicheng Zheng}
\author[1]{Lihan Zha}
\author[2]{Chunwei Xing}
\author[2]{Rajdeep Singh}
\author[2]{Omar Hossain}
\author[2]{Antonio Loquercio}
\author[1]{Dhruv Shah}

\affiliation[1]{Princeton University}
\affiliation[2]{University of Pennsylvania}
\contribution[*]{Equal contributions.}

\begin{document}

\title{\method: Long-Horizon Reasoning \& Navigation with a Visuo-Spatio-Temporal Memory}

\abstract{
    \input{secs/0-abstract}
}


\website{https://ravenmem.github.io}{ravenmem.github.io}

\code{https://github.com/princeton-prism/RAVEN}{github.com/princeton-prism/RAVEN}

\maketitle


\input{secs/1-intro}
\input{secs/2-relatedworks}
\input{secs/3-proform}
\input{secs/4-method}
\input{secs/5-experiments}
\input{secs/6-conclusion}
\input{secs/7-limitation}

\clearpage
\bibliographystyle{unsrt}
\bibliography{references.bib}

\clearpage
\beginappendix{
\input{secs/8-appendix}

}

\end{document}

%% file: preamble/notation.tex
\makeatletter
\newcommand{\longdash}[1][2em]{%
  \makebox[#1]{$\m@th\smash-\mkern-7mu\cleaders\hbox{$\mkern-2mu\smash-\mkern-2mu$}\hfill\mkern-7mu\smash-$}}
\makeatother
\newcommand{\omitskip}{\kern-\arraycolsep}

%% file: secs/0-abstract.tex


Long-term robot deployment requires a compact and scalable memory that preserves fine-grained visual semantics, grounds observations in space and time, and enables efficient storage and retrieval. In this paper, we propose \method, an agentic memory system for long-horizon robotic question answering and navigation. \method stores visual embeddings with pose and time in a vector database, and grounds retrieval in a spatial map to answer queries and navigate to goals. By operating directly on visual embeddings, \method avoids lossy image-to-text captioning and enables accurate semantic, spatial, and temporal retrieval at scale. Across several simulated and real-world video question-answering benchmarks, \method consistently surpasses caption-based memory systems and matches frontier VLMs on long-horizon tasks at 10$\times$ lower retrieval cost. Finally, we instantiate \method on a Unitree Go1 robot for the task of long-horizon navigation for natural language goal-reaching, and show successful deployment over several large indoor environments.

%% file: secs/1-intro.tex
\input{fig_inputs/fig_teaser}
\section{Introduction}

Memorizing and navigating to previously visited places is a hallmark of human spatial intelligence and essential for lifelong robot deployment. A supermarket assistant, for example, must handle queries ranging from specific product locations to fine-grained descriptions---like \emph{a blue hat with a pink bow}---requiring rich semantic and spatial information. Conventional robotic memory representations struggle with this. Categorical semantic maps with predefined labels~\citep{chang2023goatthing, maturana2017real, 8411109, 9341738, xie2025osmagllmzeroshotopenvocabularyobject} and metric maps such as 3D point clouds restrict indexing to a closed vocabulary, so a query like \emph{Where did I last see a blue hat with a pink bow?} fails the moment its distinguishing details fall outside that set. Raw video streams retain complete observations but are too large and redundant to serve as memory directly. 
Open-vocabulary language supervision~\citep{anwar2024remembrbuildingreasoninglonghorizon, mao2025metamemoryretrievingintegratingsemanticspatial} is a natural response, and most current pipelines implement it by captioning observations and storing text embeddings. This introduces a lossy image-to-text captioning stage: a \textit{captioning bottleneck} that compresses shape, spatial layout, and style into coarse text and leaves retrieval brittle whenever the discriminating detail was never captioned. This motivates a memory representation that is compact yet information-rich---and prompts the question of whether we can move beyond text-based memory toward a direct visual representation, and even something simpler.

To tackle this bottleneck, we propose using rich visual embeddings as the fundamental building block of a semantic memory module. In this work, we introduce \methodfull (\method), a memory system that operates directly on visual embeddings. By leveraging the multimodal embedding space of modern visual foundation models, \method preserves fine-grained visual semantics that are lost in text-based surrogates, while maintaining the scalability of vector databases that support efficient lookup~\cite{douze2024faiss}. 
Illustrated in Fig. \ref{fig:teaser}, RAVEN encodes egocentric video frames with a multimodal encoder and stores them in a vector database as visuo-spatio-temporal memory triplets. For retrieval, we leverage efficient Retrieval-Augmented Generation (RAG) and a tool-using vision language agent to answer user questions and guide the robot to the target by visual reasoning and planning~\cite{shinn2023reflexion,yao2022react}.
We evaluate RAVEN on a suite of challenging video-retrieval benchmarks and real-world navigation tasks. We introduce a new robot memory benchmark \ourdataset, which tests long-horizon visuo-spatio-temporal retrieval in simulated and real-world egocentric trajectories.Additionally, we validate \method on existing long-horizon retrieval benchmarks NaVQA \cite{anwar2024remembrbuildingreasoninglonghorizon} and FindingDory \cite{yadav2024findingdory}. We also demonstrate real-world feasibility via deployment on a physical robot. 
Our experiments demonstrate that \method consistently outperforms caption-based methods across 4 real-world and 19 simulated environments, as well as on standard navigation benchmarks, particularly for queries that demand fine-grained visual details. Extensive evaluations further reveal that \method scales favorably with memory context: it retains over 97\% of its performance on frontier base models and still preserves 67\% on small open-source models, while achieving over 250$\times$ storage compression without accuracy loss, and 10$\times$ greater retrieval efficiency compared to a na\"ive VLM.

%% file: fig_inputs/fig_teaser.tex
\begin{figure*}[h!]
  \centering
  \includegraphics[width=\linewidth]{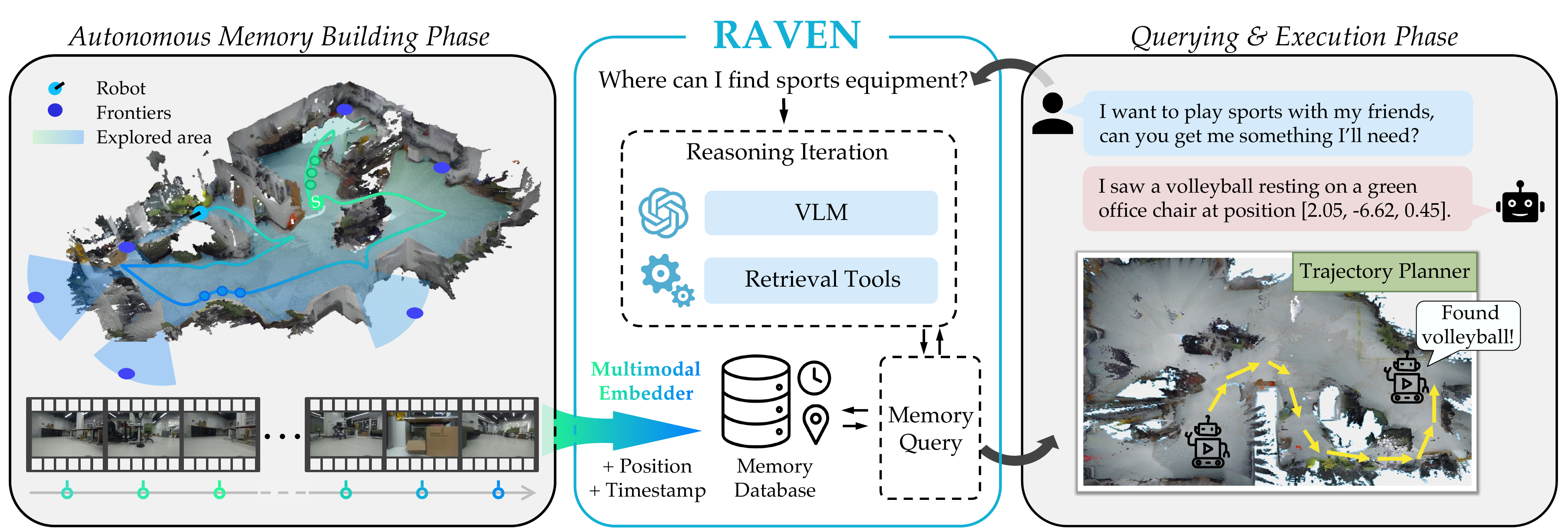}
  \caption{\method leverages visual embeddings as a long-term memory for robotic question answering and navigation. With multimodal embedding models, \method{} computes frame- and segment-level visual embeddings and stores them with corresponding poses and timestamps in a memory database. At query time, a VLM agent iteratively invokes function calls to retrieve top-$k$ relevant memory entries and generates answers. The robot can further guide the user to the predicted answer location by navigation with a trajectory planner.}
  \label{fig:teaser}
\end{figure*}


%% file: secs/2-relatedworks.tex
\section{Related Work}

Navigating large environments is fundamentally a partially observable problem, necessitating a memory system to maintain global context of the environment. Early approaches towards building semantic memory relied on explicit memory~\cite{KOSTAVELIS201586,wolf2008semantic}, attaching semantic labels to metric maps such as point clouds~\cite{nuchter2008towards, seichter2022efficient,chaplot2020object, 8411109, chang2023goatthing}, occupancy grids~\cite{sunderhauf2016place,9341738,maturana2017real}, or topological graphs~\cite{galindo2008robot, lang2014semantic}. While interpretable, these methods suffer from a limited vocabulary; encoding memory with a small number of discrete concepts discards fine-grained spatial and relational details, causing out-of-vocabulary information to vanish. To address this, recent work has shifted toward implicit cognitive memory, which utilizes neural representations to encode scenes and attributes~\cite{dorbala2022clip,liu2024cognitive,anwar2024remembrbuildingreasoninglonghorizon,mao2025metamemoryretrievingintegratingsemanticspatial}. However, current state-of-the-art implicit pipelines introduce new bottlenecks. Methods like ReMEmbR~\cite{anwar2024remembrbuildingreasoninglonghorizon} rely on explicit captioning, introducing a lossy image-to-text conversion that overlooks visual details difficult to verbalize, such as texture or precise spatial relations (see Appendix~\ref{appx_short:visual_and_examples}). Conversely, approaches like CLIP-Fields~\cite{_Mahi_Shafiullah_2023} avoid this language bottleneck by grounding features in a learned 3D field, but their reliance on per-scene optimization limits scalability across changing environments. Similarly, RNN-based memories~\cite{yang2025improving} struggle with long-term forgetting, while retrieval-free context models~\cite{wang2024navformer,sridhar2025memer} fail to scale to extremely long sequences.

To overcome the limitations of captioning bottlenecks and per-scene training, we revisit the potential of raw visual embeddings as a scalable memory substrate. While early embedding approaches suffered from degradation when derived from object detectors~\cite{yang20253dmem3dscenememory} or dimensionality reduction~\cite{li2025episodic}, modern visual representations have increasingly aligned with text driven by scaling laws~\cite{kaplan2020scaling, zhai2023sigmoidlosslanguageimage,seed16embedding}. Unlike methods that embed 3D data~\cite{hu20253dllm}—which are often lossy or scale-limited—we utilize dense visual embeddings that capture holistic scene semantics without intermediate textual compression. Recent analyses confirm that RAG systems benefit significantly from retrieving with multimodal embeddings rather than text surrogates~\cite{lumer2025comparisontextbasedimagebasedretrieval}. Motivated by this, \method operates directly on compact visual embeddings, retaining maximal information density while ensuring the generalization required for open-vocabulary tasks.


The retrieval mechanism is equally critical for long-horizon operation. Scaled dot-product attention for memory fusion~\cite{hu20253dllm,wang2021temporal} requires training downstream LLMs and scales linearly or quadratically per query. \method instead uses conventional vector retrieval~\cite{douze2024faiss}, which is robustly optimized toward sub-linear complexity and lets us scale to days of experience without paying the cost of full-context processing. By combining the high-fidelity information of visual embeddings with the efficiency of vector databases, we propose a memory system that supports complex, agentic reasoning loops while remaining computationally lightweight and training-free.

%% file: secs/3-proform.tex
\section{Setup and Overview}


We frame robotic question answering as a two-phase navigation task: given a query, the robot first builds a memory of the environment; during deployment, the robot plans and navigates using only this memory, without any additional exploration. We study a set of queries that commonly arise in real-world robot navigation, spanning fine-grained visual attributes, temporal references, location-centric questions, abstract concepts tied to a place, multi-hop reasoning, and on-object text to extract.

\textbf{Phase I: Building Memory through Exploration} 
\label{sec:setup_phase_1}
In this phase, the robot is deployed to actively explore the environment over a fixed time horizon, building a map (e.g., by running SLAM) and storing its perceptions. The robot acquires a sequence of observations in the form of RGB video frames $O_{1:N}$, depth maps, and proprioceptive data (e.g., pose, location) with timestamps. It will process the observation data and store intermediate results in its memory for future use. For convenience, we denote the collected RGB video frames ($H\times W\times C$ arrays) as $\{o_i | o_i \in \mathbb{R}^{H\times W\times C}, i = 1, 2, 3, \ldots, N\}$, and the 3D position as $\{p_i |  p_i  =  (x_i, y_i, z_i)\in \mathbb{R}^3, i=1,2, 3, \ldots, N\}$.

\textbf{Phase II: Planning and Navigating with Memory} 
\label{sec:setup_phase_2}
Let $\mathcal{Q}$ denote the query dataset. In the second phase, the robot addresses the natural language query $q \in \mathcal{Q}$ about something in the previously observed environment. It will parse the query and retrieve relevant memories of the target in its pre-built memory database in Phase I. Once the robot recalls the location of the query $q$, it directly navigates to it via the shortest reachable path (e.g., the $A^*$ method~\cite{4082128}), without any on-the-fly exploration or in-situ trial-and-error. A navigation episode is considered successful if the robot arrives within a tolerable distance of the ground-truth target(s). We calculate the task success rate (SR). If the Phase 1 exploration fails to capture an object, Phase 2 cannot retrieve it, and the query results in failure. In addition, we investigate question-answering (QA) tasks grounded in navigation and planning as an internal justification mechanism for planning.

%% file: secs/4-method.tex
\section{RAVEN: A Semantic Memory System for Long-Horizon Navigation}
\label{sec:method}

\input{fig_inputs/fig_pipeline}
We present \method, an elegant memory architecture illlustrated in Fig.~\ref{fig:system}. During the \emph{Memory Building} stage, the memory retrieval module is constructed and packaged as a tool for agent use (\ref{sec:method_raven}); during the \emph{Memory Querying} stage, the overall design leverages an agentic loop that iteratively performs perception, reasoning, and tool usage as a state machine (\ref{sec:method_agent}). Given an input query, the system produces either a direct answer or a target location $\hat{p} = (\hat{x}, \hat{y}, \hat{z})$ for downstream planning and navigation (\ref{sec:execute}). 

\vspace{2pt}
\subsection{Retrieving from Visuo-Spatio-Temporal Memories}
\label{sec:method_raven}
To enable grounding in space and time, visual embeddings alone are insufficient. Therefore, we construct a composite memory entry for every observed frame. As the robot explores, we encode RGB frames $o_i$ directly into compact latent representations $z_i$ using a pretrained multimodal encoder (e.g., CLIP~\cite{radford2021learningtransferablevisualmodels}, SigLIP~\cite{zhai2023sigmoidlosslanguageimage}, Seed1.6-Embedding~\cite{seed16embedding}, and QQMM-v2~\cite{xue2025improvemultimodalembeddinglearning}). These embeddings are indexed alongside their corresponding robot pose $p_i$ and world-clock timestamp $t_i$ in a vector database. This triplet memory structure design (named \textbf{visuo-spatio-temporal memory}) allows for flexible retrieval queries based on semantic similarity, spatial proximity, or temporal windows, denoted as $\{m_i\}_N := \{(p_1, t_1, z_1,o_1), (p_2, t_2,z_2, o_2), \ldots, (p_N, t_N, z_N, o_N)\}$.

Memory retrieval is exposed to the agent as a callable tool (\ref{sec:method_agent}). It is supported by standard retrieval frameworks, e.g., FAISS~\cite{johnson2019billion} or Milvus~\cite{10.1145/3448016.3457550}, which enable efficient nearest-neighbor search. Given a query—either a scalar or a vector—the retrieval system returns the top-$K$ items in the database under a chosen distance metric (e.g., Euclidean or cosine distance). These systems are optimized for large-scale datasets, enabling retrieval over an $N$-sized database with sub-$O(N)$ time complexity.

\vspace{2pt}
\subsection{Reasoning with Tool-Using Agent}
\label{sec:method_agent}

At the \emph{Query} stage, the robot is given a question $q$ (e.g., \emph{Where can I play some sports with my friends} in Fig.~\ref{fig:system}). Rather than performing a single retrieval and directly returning the matched result, \method aims to improve retrieval precision by leveraging a strong vision-language agent. Specifically, a VLM perceives retrieved visual memories, reasons over the contextual information, and orchestrates actions iteratively in a loop, formulated as a finite state machine. 

It maintains a current context $R_0$ as its working memory and determines whether the accumulated evidence is sufficient to produce a final answer. If not, it invokes a retrieval action. In \method, four primitive tools are available: \textbf{text-based} (comparing an agent-generated query $\hat{q}$ against stored visual memories $z_i$ via cosine distance $1 - \langle f_{\text{enc}}(\hat{q}), z_i \rangle$), \textbf{time-based} (returning consecutive memories starting from a specific timestamp), \textbf{position-based} (finding spatial nearest-neighbors to a proposed location $(\hat{x}_q, \hat{y}_q, \hat{z}_q)$), and \textbf{image-based} retrieval (we uncover the potential to extend user queries to images in Appendix~\ref{appx_short:results_and_implement_details}). 

Each retrieval call returns $K$ memory entries $\{m_i'\}_K$, each containing the retrieved location, timestamp, and image, which are appended to the context at reasoning step $\tau$, yielding $R_{\tau+1} = R_{\tau} \cup \{m_i'\}_K$. The VLM evaluates the updated working memory $R_{\tau+1}$ to determine if the accumulated evidence is sufficient. In this way, \method forms a closed-loop reasoning process over visual memory, effectively shortening the attention window presented to the agent and avoiding costly exhaustive searches over all trajectories. Once confident, the VLM exits the loop and produces a structured output: (i) a textual answer, (ii) a spatial goal $(\hat{x},\hat{y},\hat{z})$, or (iii) a temporal reference. We defer detailed tool implementations, hyperparameter selection, and step-by-step examples to Appendices~\ref{appx_short:results_and_implement_details} and \ref{appx_short:visual_and_examples}.
Overall, our \method features a general visual memory system and a VLM agent reasoning mechanism that is highly effective in real-world deployments, yet simple, efficient, and training-free. These merits make \method a plug-and-play memory system, scalable to extra-long trajectories and practical for embodied decision-making.

\vspace{2pt}
\subsection{Exploration \& Navigation Execution}

\label{sec:execute}

\method can be seamlessly integrated into general robot planning, control, and navigation tasks. We deploy a Unitree Go1 quadruped robot with a ZED 2i camera and a Jetson Orin NX 16GB, as demonstrated in Fig.~\ref{fig:robot}.

\textbf{Exploration} At the beginning of deployment, the robot actively explores the environment (e.g., via greedy frontier exploration~\cite{yokoyama2024vlfm}) while continuously collecting multimodal observations (e.g., RGB images $\{o_i\}_{i=1}^N$), together with depth maps and other associated robot states, including spatial poses $p_i = (x_i, y_i, z_i)$ and timestamps $t_i$. We use RGB-D SLAM to reconstruct a 3D occupancy map, as shown in Fig.~\ref{fig:teaser} (left).

\textbf{Navigation} During navigation execution, predicted spatial coordinates are passed to an $A^*$~\cite{4082128} trajectory planner, closing the retrieval--planning loop and enabling real-world deployment, as shown in the right panel of Fig.~\ref{fig:teaser}. Implementation details are provided in Appendix~\ref{appx_short:results_and_implement_details}.

\input{tabs/real_world_high_level}
\input{tabs/findingdory_table}

%% file: fig_inputs/fig_pipeline.tex
\begin{figure*}[t]
    \centering
    \includegraphics[width=1.0\linewidth]{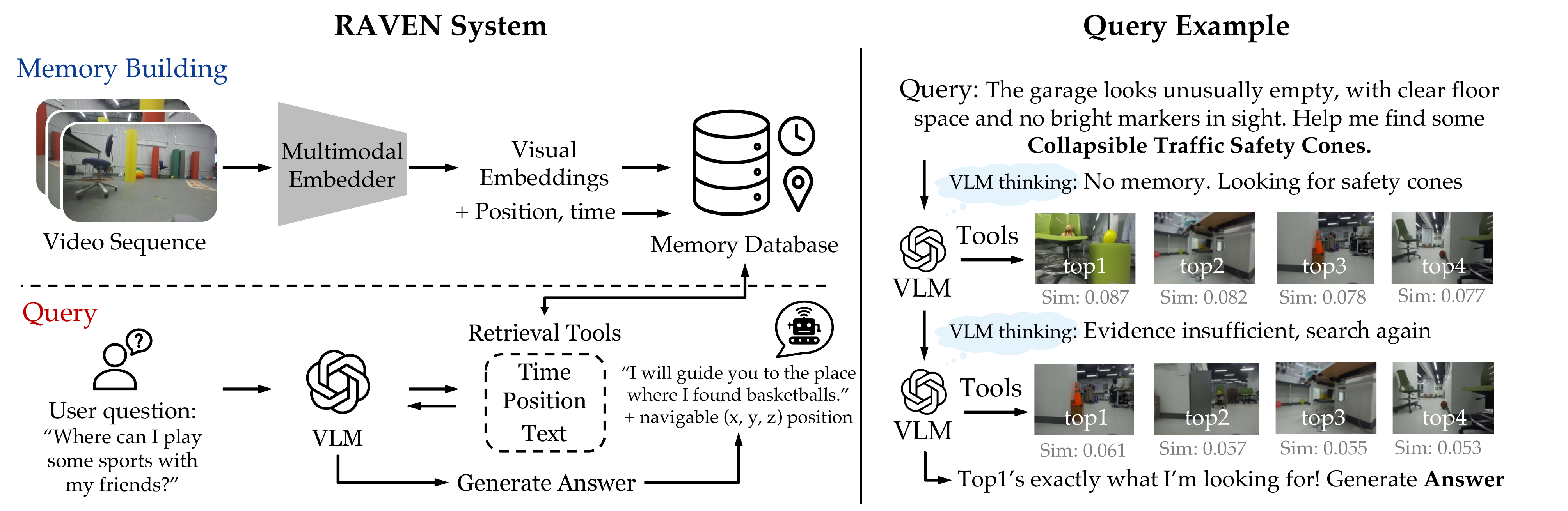}
    \caption{Overview of \method pipeline and evaluation setup. (Left) We design RAVEN with a visual memory building phase and a querying phase. The memory building phase runs on multimodal embedders, embeds the frames, then stores the image embedding, position and time vectors into a vector database. Then, when a user asks a question, a vector database querying loop starts with an VLM and gives answer for the query. (Right) Query example about safety cones and reasoning example when a VLM agent uses function calls for memory retrieving several times until it feels confident to generate answers.}
    
    \label{fig:system}
\end{figure*}

%% file: tabs/real_world_high_level.tex
\begin{table*}[t]
\centering
\caption{Overall system accuracy on the robot-egos splits of \ourdataset, including real-world and simulation scenes. Visual embedders with closed models outperform text-based or VLM-based memory systems in our real-world robot experiments.}
\label{tab:real_and_sim_main_acc_result}

\footnotesize
\setlength{\tabcolsep}{1.0pt}
\renewcommand{\arraystretch}{1.0}

\begin{adjustbox}{max width=\textwidth}
\begin{tabular}{llccc cc c}
\toprule
\multicolumn{2}{c}{} &
\multicolumn{3}{c}{\textbf{Closed VLMs}} &
\multicolumn{2}{c}{\textbf{Open VLMs}} &
\multicolumn{1}{c}{Embedder} \\
\cmidrule(lr){3-5}\cmidrule(lr){6-7}
\multicolumn{2}{c}{Mean$\pm$Std} &
Gemini-2.5-Flash & GPT-5.2 & Gemini-3-Pro &
Gemma3-27b & Qwen3-VL-32b &
Only \\
\midrule
\midrule
\multicolumn{8}{c}{\textbf{Real-World Robot Tasks} (Scene 1$\sim$4, 10 seeds, 21 queries)} \\
\midrule

\multirow{3}{*}{\textbf{Embedders}}
& QQMM-v2
 & \textbf{92.38$\pm$3.33} & 93.33$\pm$3.33 & 90.00$\pm$3.51
 & \textbf{67.62$\pm$3.76} & 30.00$\pm$5.52
 & \textbf{66.67} \\
& Seed (Closed)
 & 85.71$\pm$5.02 & 93.33$\pm$3.33 & \textbf{90.47$\pm$4.49}
 & 45.71$\pm$6.43 & 31.90$\pm$5.96
 & 66.67 \\
& SigLIP
 & 84.76$\pm$3.01 & \textbf{97.14$\pm$2.46} & 90.00$\pm$4.17
 & 47.62$\pm$4.49 & \textbf{35.24$\pm$6.43}
 & 42.86 \\
\cmidrule{1-8}
& \vlmonly
 & 85.24$\pm$5.70 & 92.86$\pm$5.14 & 86.19$\pm$1.51
 & 14.29$\pm$4.49 & 4.76$\pm$3.89
 & N/A \\
& ReMEmbR (QQMM-v2)
 & 76.67$\pm$3.51 & 74.76$\pm$3.92 & 80.48$\pm$3.51
 & 14.76$\pm$6.53 & 28.57$\pm$6.35
 & N/A \\

\midrule
\midrule
\multicolumn{8}{c}{\textbf{Habitat Simulation Tasks} (Scene 1$\sim$19, 3 seeds, 157 queries)} \\
\midrule

\multirow{3}{*}{\textbf{Embedders}}
& QQMM-v2
 & 80.47$\pm$0.37 & 75.58$\pm$1.95 & 86.62$\pm$0.00
 & 54.99$\pm$3.51 & 26.11$\pm$2.21
 & 61.78 \\
& Seed (Closed)
 & \textbf{81.74$\pm$0.37} & 76.22$\pm$1.60 & 86.84$\pm$0.74
 & \textbf{60.30$\pm$0.97} & \textbf{27.39$\pm$0.64}
 & \textbf{62.42} \\
& SigLIP
 & 81.10$\pm$2.24 & 71.97$\pm$0.64 & 86.20$\pm$2.05
 & 56.05$\pm$2.30 & 22.51$\pm$3.89
 & 61.78 \\
\cmidrule{1-8}
& \vlmonly
 & 76.65$\pm$1.33 & \textbf{80.47$\pm$2.05} & \textbf{88.75$\pm$1.33}
 & 11.04$\pm$0.97 & 9.77$\pm$0.37
 & N/A \\
& ReMEmbR (QQMM-v2)
 & 54.99$\pm$3.51 & 57.32$\pm$1.27 & 61.36$\pm$0.37
 & 15.71$\pm$4.10 & 20.38$\pm$1.10
 & N/A \\

\bottomrule
\end{tabular}
\end{adjustbox}
\end{table*}

%% file: tabs/findingdory_table.tex
\begin{table*}[t]
\centering
\caption{High-level success rate on the FindingDory benchmark~\cite{yadav2024findingdory}. \method using the QQMM-v2 visual embeddings consistently outperform ReMEmbR and can often outperform state-of-the-art foundation models (VLM Only). \method is significantly more efficient than querying a VLM directly, enabling scaling up to long-term deployments in large environments.}
\label{tab:findingdory_results}
\footnotesize
\setlength{\tabcolsep}{3pt}
\renewcommand{\arraystretch}{1.08}
\resizebox{\textwidth}{!}{
\begin{tabular}{ccc lc c cc cc c}
\toprule 
& Avg \# of Frames & &
&
Ours &
\multicolumn{2}{c}{ReMEmbR (Caption)} &
\multicolumn{3}{c}{Other Baselines}
 \\
\cmidrule(lr){5-5}
\cmidrule(lr){6-7}
\cmidrule(lr){8-10}

\#Videos & per Video & \#Queries &
Statistics\% & QQMM-v2 &
QQMM-v2 & MixedBread &
\vlmonly & Embedder Only & Random
\\
\midrule
\midrule

\multicolumn{10}{c}{\textbf{The Long-Horizon Subset ($ep\_91$ to $ep\_100$)}} \\

\midrule
\multirow{5}{*}{10}
& \multirow{5}{*}{3022.9}
& \multirow{5}{*}{596} & Overall Accuracy & \textbf{32.2} & 25.0 & 25.2 & 28.2 & 18.8 & 9.2 \\
& &  & Single Spatial & \textbf{45.1} & 36.7 & 36.0 & 36.0 & 31.0 & 13.4 \\
& &  & Single Temporal& 24.8 & 17.1 & 18.6 & \textbf{28.1} & 8.1 & 6.9 \\
& &  & Multi-Goal Navigation & \textbf{6.7} & 4.5 & 4.5 & 2.2 & 3.4 & 0.5 \\
\cmidrule{4-10}
& &  & $\alpha$\% (Memory Retrieved) & \textbf{7.43\%} & 11.17\% & 10.68\% & 100\% & 0.0\% & N/A \\
\midrule
\midrule

\multicolumn{10}{c}{\textbf{The Full Dataset ($ep\_1$ to $ep\_100$)}} \\

\midrule

\multirow{5}{*}{100}
& \multirow{5}{*}{1794.2}
& \multirow{5}{*}{5870} & Overall Accuracy & 35.1 & 27.2 & 27.3 & \textbf{38.2} & 21.5 & 11.1 \\
& &  & Single Spatial & \textbf{46.3} & 37.7 & 37.5 & 45.1 & 32.8 & 15.1 \\
& &  & Single Temporal & 30.1 & 21.9 & 22.7 & \textbf{38.9} & 12.8 & 9.6 \\
& &  & Multi-Goal Navigation & 9.4 & 4.7 & 4.1 & \textbf{13.6} & 4.0 & 1.2 \\
\cmidrule{4-10}
& &  &  $\alpha$\% (Memory Retrieved) & \textbf{10.0\%} & 16.5\% & 16.4\% & 100.0\% & 0.1\% & N/A \\

\bottomrule
\end{tabular}
}
\end{table*}

%% file: secs/5-experiments.tex

\input{fig_inputs/fig_performance}

\input{fig_inputs/fig_rollouts}
\section{Experiments}
\label{sec:experiments}



Our evaluation addresses three questions: (i) Information-Richness: Do raw visual embeddings recover fine details better than captions? (ii) Scalability: Can \method maintain retrieval accuracy as the memory horizon scales to thousands of frames? (iii) Real-World Feasibility: Does the system transfer effectively to physical robots? We tailor the evaluation suites (\ref{sec:experiments_eval}) and baseline candidates (\ref{sec:experiments_baselines}) to address these questions (\ref{sec:experiments_main_results}).

\vspace{2pt}
\subsection{Memory Benchmarks}
\label{sec:experiments_eval}
We compare \method against strong baselines on \navqa~\cite{anwar2024remembrbuildingreasoninglonghorizon}, \findingdory~\cite{yadav2024findingdory}, and a self-curated, more challenging navigation QA benchmark \ourdataset, and provide more details in Appendix~\ref{appendix:dataset}.

\textbf{\navqa}~\cite{anwar2024remembrbuildingreasoninglonghorizon}. A video question answering dataset built on top of the CODa robot navigation dataset \cite{zhang2023robustrobot3dperception}, which was collected in real-world urban scenes. The dataset covers basic descriptive, spatial, and temporal queries across short-, medium-, and long-horizon episodes.

\textbf{\findingdory} \cite{yadav2024findingdory}. A memory-based single- or multi-goal navigation benchmark based on the Habitat simulation~\cite{habitat19iccv}. In this benchmark, the robot performs spatiotemporal navigation tasks by retrieving memories from loco-manipulation history episodes, in which it picks up and places down various objects at different locations in a specified order.

\textbf{\ourdataset}. To enable a more comprehensive evaluation of robotic memory retrieval under challenging settings and complex queries, we curate a customized dataset, \ourdataset, designed to assess advanced retrieval capabilities from real-world, simulated, and web-sourced tour videos. We intentionally categorize queries into diverse types, including \emph{dominant object}, \emph{secondary object}, \emph{reasoning-based}, \emph{information recall}, and \emph{spatial understanding}. Details and statistics of \ourdataset are provided in Appendix~\ref{appendix:dataset}.

\subsection{Metrics}
For \navqa, we report descriptive accuracy, along with positional and temporal error statistics, as well as the overall accuracy. 
For \findingdory, we present the high-level success rate on the validation set. To further demonstrate the long-horizon efficacy, we focus on extra-long episodes from $ep\_91$ to $ep\_100$, each containing on average 3{,}000 frames. Given the large memory scale, we further compute $\alpha$, the fraction of memory items retrieved for each query, and use its inverse as a measure of memory efficiency $E$, denoted as 
$\alpha (\%):=\frac{N_{\text{memory}}^{\text{retrieved}}}{N_{\text{memory}}} \times 100\%$, where $ E:=\frac{1}{\alpha(\%)} $.
For the web-sourced, self-recorded, and simulated subsets of \ourdataset, we report target retrieval accuracy on both the \emph{simple} and \emph{hard} splits. For real-world deployments, we evaluate navigation performance using task success rate. In addition, more fine-grained per-category accuracies (covering categories such as \emph{reasoning}, \emph{dominant}, and \emph{secondary}) are provided in the Appendix.

\subsection{Baselines}
\label{sec:experiments_baselines}

We compare \method against the following baselines:

\textbf{\remembr}\cite{anwar2024remembrbuildingreasoninglonghorizon}. \remembr is a representative state-of-the-art approach within the scope of memory-based navigation problem. It converts visual observations into textual captions as memory entries using VLM and performs retrieval via text embeddings.

\textbf{\vlmonly}. We directly feed all contextual image frames to a VLM, prompting it to predict the target in a single forward pass. When the number of frames exceeds the context window, we first subsample the frames. The method involves neither explicit retrieval nor an agentic reasoning loop. 

\textbf{Embedder Only}. This method embeds images using multimodal embedding models and returns the image whose embedding has the highest similarity to the user query embedding. No language model is used in this process.


\subsection{Real-world Results}
\label{sec:experiments_main_results}

We test RAVEN against state-of-the-art baselines on real-world indoor navigation tasks. We summarize results in Fig.~\ref{fig:performance_hist}, and direct the reader to Appendix~\ref{appx_short:results_and_implement_details} and \ref{appx_short:visual_and_examples} for a more detailed discussion.

\textbf{\method consistently outperforms the state-of-the-art (\remembr).} As shown in Tables~\ref{tab:real_and_sim_main_acc_result}, \ref{tab:findingdory_results}, and \ref{tab:IRS_simple_hard}, \method substantially surpasses the caption-based baseline, achieving up to 95.1\% accuracy on hard queries (Seed1.6-Embed + Gemini-3-Pro) and 92.7\% with QQMM-v2 + Gemini-3-Pro (Table~\ref{tab:IRS_simple_hard}).
Notably, \remembr frequently fails due to either under-captioning (losing visual details due to \emph{captioning bottleneck}) or over-captioning (introducing irrelevant noise or misleading decriptions) (Appendix~\ref{appx_short:visual_and_examples}).
Furthermore, the performance gap widens from $\sim$13\% on Simple queries to 30\% on Hard queries. This confirms that direct visual embeddings robustly preserve fine-grained semantics (e.g., secondary objects) where the captioning bottleneck fails.


\input{fig_inputs/fig_compact}
\textbf{\method outperforms and scales more efficiently than frontier VLMs.} In Table~\ref{tab:IRS_simple_hard}, for open VLMs, there is a sharp performance drop ($\sim$13\%) on \vlmonly, while open-VLM \method still maintains comparable ($\sim$90\%) to closed-VLM ones by pruning redundant memory context. On the closed model side (Table~\ref{tab:findingdory_results}), \method surpasses \vlmonly by 4\% on the longest-horizon subset (up to 3,022 frames). Notably, \method's memory efficiency $E$ is 10$\times$ higher (Fig.~\ref{fig:performance_hist}). Unlike \vlmonly, \method's precise retrieval shortens working memory, allowing the VLM to allocate undiluted attention to relevant frames (Fig.~\ref{fig:real_robot_rollouts}).
This advantage stems from \method's highly optimized compactness (Fig.~\ref{fig:high_compactness}), which preserves 90\% of capability even when videos are downsampled to $0.12$ fps, yielding a $22{,}315\times$ compression rate over raw RGB data.

\textbf{Medium-sized open VLMs underperform embedder-only baselines.} As shown in Tables~\ref{tab:real_and_sim_main_acc_result}~\ref{tab:IRS_simple_hard}, when using 27B$\sim$32B open VLMs, \method, \remembr, and \vlmonly, all perform worse than or only slightly better than the multimodal embedder-only baseline. Therefore, for offline deployments without Internet access, directly adopting an embedder-only approach may be preferable to incorporating medium-sized open VLMs.

\textbf{Real-world deployment of \method.}
We report qualitative deployment results in Fig.~\ref{fig:real_robot_rollouts}.
Despite the presence of challenging queries involving multi-step reasoning, small object retrieval, and fine-grained visual information recall, \method achieves up to 97.1\% success rate (SigLIP + GPT-5.2) and 92.4\% with our main configuration (QQMM-v2 + Gemini-2.5-Flash), substantially outperforming alternative approaches.


\textbf{Synergy between visual and visuo-spatio-temporal memory.} As demonstrated in Table~\ref{tab:navqa_results}, utilizing images rather than text as VLM inputs enhances visuo-spatio-temporal retrieval performance, which implies their mutual benefits.


\label{sec:experiments_real_world}

%% file: fig_inputs/fig_performance.tex
\begin{figure}[t]
    \centering
    \footnotesize
    \includegraphics[width=1.0\linewidth]{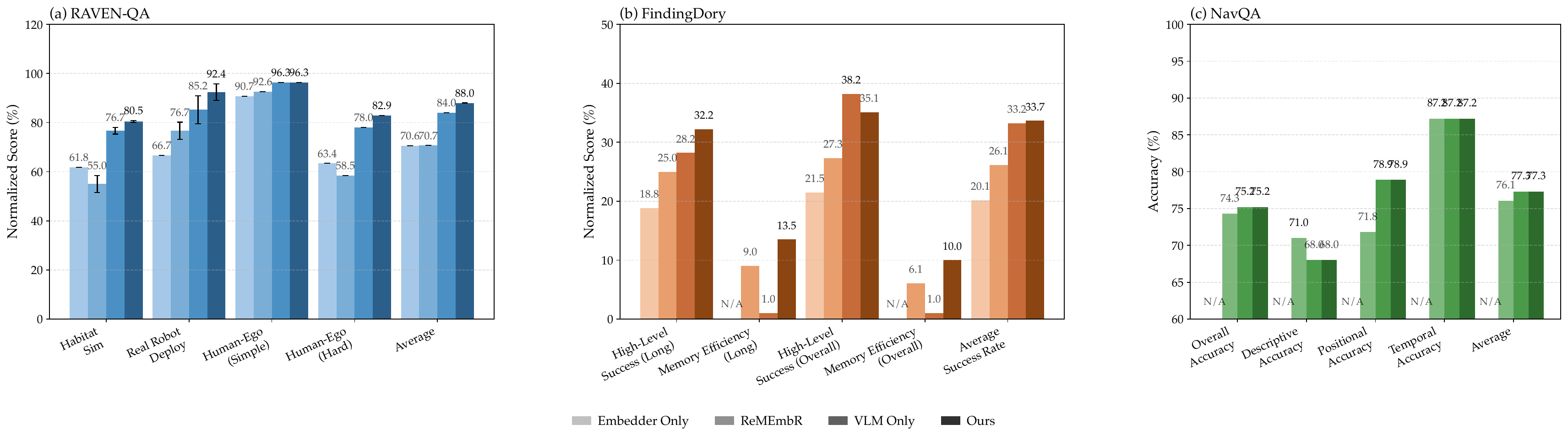}
    \caption{Performance on \findingdory, our \ourdataset and \navqa. Results on \findingdory and \ourdataset are evaluated using Gemini-2.5-Flash~\cite{comanici2025gemini} as the VLM and QQMM-v2
    \cite{xue2025improvemultimodalembeddinglearning} as the embedder. Results on \navqa are evaluated using Gemini-3-Pro~\cite{gemini-3-pro} and QQMM-v2~\cite{xue2025improvemultimodalembeddinglearning}. Error bars are the standard deviations across multiple runs.}
    \label{fig:performance_hist}
\end{figure}

%% file: fig_inputs/fig_rollouts.tex
\begin{figure*}[ht]
  \centering
  \includegraphics[width=1.004\linewidth]{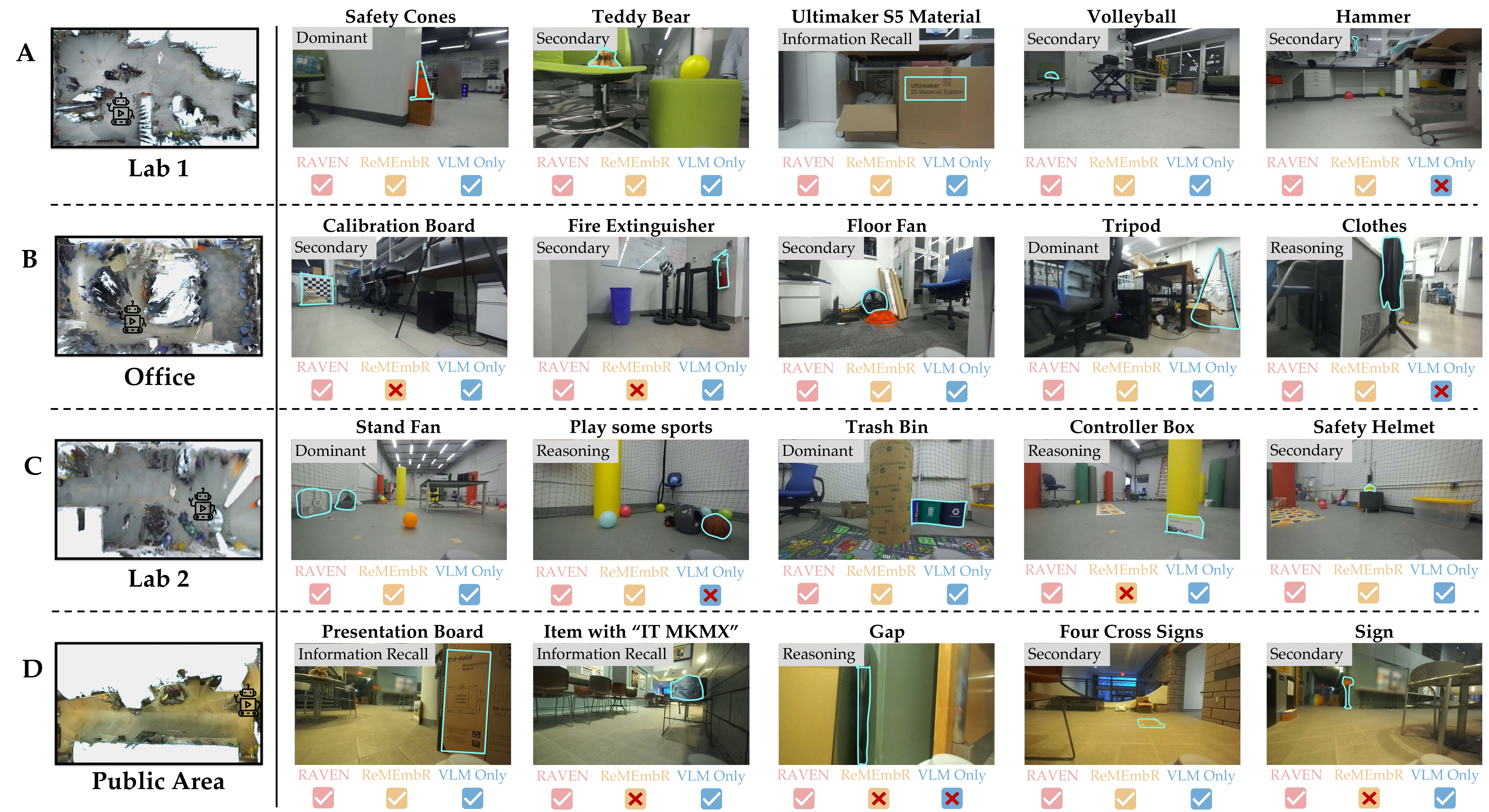}
  \caption{Rollouts from real-world robot deployment. (Left) Birds-eye view reconstruction map of four scenes (not available to \method). (Right) For each method, success and failure cases are shown in check and cross marks, respectively. We provide the ground-truth frames that locate the requested targets, along with the original queries and their categories. Detailed reasoning steps and analyses of \method, \remembr, and \vlmonly on the real-world robot split are provided in the Appendix~\ref{appx_short:visual_and_examples}.}
  \label{fig:real_robot_rollouts}
\end{figure*}

%% file: fig_inputs/fig_compact.tex
\begin{wrapfigure}{r}{0.55\textwidth}
    \vspace{-10pt}
    \centering
    \includegraphics[width=\linewidth]{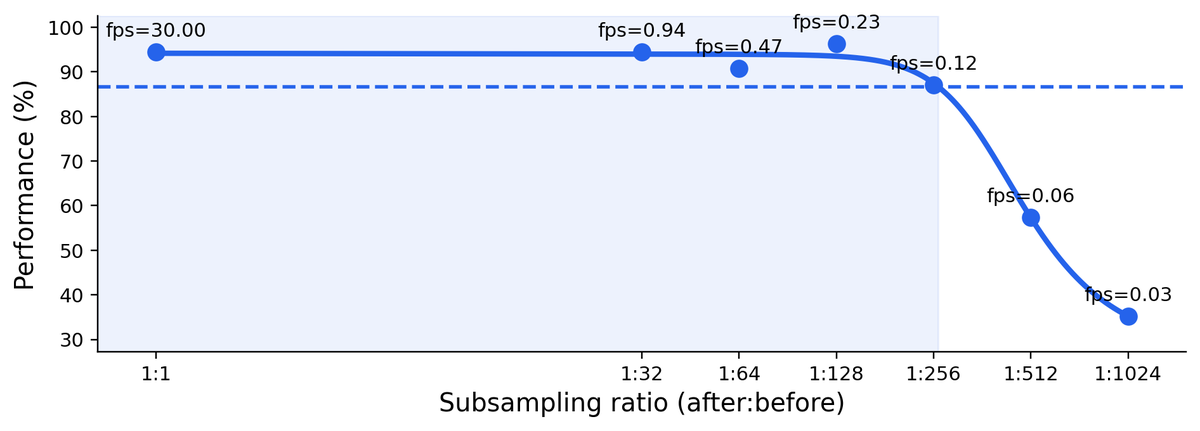}
    \caption{RAVEN is highly compact and scalable. Dashed line shows $10\%$ performance drop at $250\times$ compression.}
    \label{fig:high_compactness}
    \vspace{-10pt}
\end{wrapfigure}




%% file: secs/6-conclusion.tex
\section{Conclusion}
\label{sec:conclusion}

We presented \method, an agentic memory system that uses a vector bank of multimodal embeddings  for long-horizon robot navigation. By constructing a visuo-spatio-temporal memory, our approach preserves fine-grained semantic details that are often lost in text-based surrogates, enabling more accurate retrieval for open-vocabulary queries. Our experiments demonstrate that \method consistently outperforms caption-based baselines and \vlmonly approaches, particularly on tasks requiring the recall of secondary objects and subtle visual features. The broader implication of our findings is in how we can leverage multimodal foundation models for robotic memory:
rather than compressing visual experience into language immediately, or compressing visual history into a prmopt, retaining high-fidelity sparse visual representations along with retrieval allows agents to reason about the world with greater nuance and precision. This validates visual embeddings as a compact, scalable, and information-rich substrate for long-horizon deployment and lifelong learning.

\noindent \textbf{Limitations:}
While \method demonstrates robust performance, it relies on the quality of pre-trained multimodal encoders; consequently, retrieval is bounded by the alignment capabilities of this embedding model. Furthermore, our current implementation treats memory primarily as a static database of past observations. While effective for stable environments, handling highly dynamic scenes where objects frequently change state or position remains an open challenge. Future work could address these limitations by incorporating domain-specific fine-tuning to better align embeddings with robotic tasks or by developing mechanisms to update and prune memory entries in response to environmental changes.

\section*{Acknowledgments}
This research was partially supported by ARL DCIST CRA W911NF-17-2-0181, DARPA TIAMAT HR0011-24-9-0430, and FrodoBots, with compute support from Google TPU Research Cloud, NVIDIA Academic Grant Program, and Gemini Academic Program.
The authors would like to thank the MIT LL and ARL T\&E Teams, and Carlos Nieto, for inspiring the problem formulation and contributing some of the simulation infrastructure that was used in this project.


%% file: secs/7-limitation.tex

%% file: secs/8-appendix.tex


\startcontents[appendices]
\printcontents[appendices]{}{1}{} 

\vspace{1cm}

\newpage
\onecolumn
\appendix

\input{secs/8.A-appendix}

\input{secs/8.B-appendix}

\input{secs/8.C-appendix}

%% file: secs/8.A-appendix.tex
\section{Dataset Details}
\label{appx_short:dataset_details}
\label{appendix:dataset}

\subsection{Overview}
\input{fig_inputs/appx_figs_input/fig_three_dataset_rollout}
Example rollout images for \navqa, \findingdory and \ourdataset are shown in Fig.~\ref{fig:three_dataset_rollout}.

\vspace{10pt}

\subsection{Details of \ourdataset}
We curated a new question-answering dataset to benchmark long-term memory retrieval in comprehensive scenarios, across simulated environments, real robot-view explorations, as well as web-sourced or self-taken indoor and outdoor tour videos. We also intentionally annotate each query with a specific category, such as reasoning, secondary object retrieval, etc. Fig.~\ref{fig:dataset} and Table~\ref{tab:raven_qa_statistics} intuitively illustrate the dataset's statistics and question distributions.

\textbf{Data Sources.} 
Specifically, \ourdataset is curated from the following sources.
\begin{itemize}
    \item \textbf{YouTube Tour Videos:} We crawl indoor and outdoor tour videos from the web (e.g., YouTube). The indoor videos consist of virtual house tours for rental listings, featuring a wide range of household items, while the outdoor videos comprise city-scale walking tours, where pedestrians in diverse outfits appear. These videos are paired with hand-crafted VQA questions, which are further divided into \emph{simple} and \emph{hard} subsets to enable more fine-grained evaluation.
    
    \item \textbf{Self-Recorded Human Videos:} We collected indoor human-perspective tour videos within our research complex. Unlike tour videos on the web, these hand-held, self-recorded videos exhibit natural camera jitter, reflecting real robot situations. They are in the \emph{hard} subset.
    
    \item \textbf{Habitat Simulation Episodes:} Robot-centric room exploration episodes are collected from 19 different scenes or initializations in Habitat simulation environments~\cite{habitat19iccv, szot2021habitat, puig2023habitat3}.
    
    \item \textbf{Real-World Robot Explorations:} To improve the benchmark's reliability for real-world transfer, we additionally include data collected from physical robot deployments. We deploy a quadruped robot (Fig.~\ref{fig:robot}) and a wheeled mobile robot (Earthrover Mini~\cite{frodobots_earth_rover_mini}). Robot-centric data are collected across more than four distinct environments, including multiple laboratories and public academic buildings. Representative rollouts are shown in Fig.~\ref{fig:real_robot_rollouts}.
\end{itemize}





\textbf{Question Taxonomy.} We categorize questions into the following types, reflecting increasing levels of perceptual, semantic, and spatial reasoning difficulty:
\begin{itemize}
  \item \textbf{Dominant:} Questions about large, salient objects that are are visually prominent but not always easy to localize due to scale and viewpoint variation.
  \item \textbf{Secondary} Questions involving small or less salient objects that require fine-grained visual memory.
  \item \textbf{Information Recall:} Fine-grained recall questions requiring reading or remembering small numbers, text, or subtle visual details.
  \item \textbf{Reasoning:} Implicit reasoning questions that do not directly name the target object, but instead require semantic inference (e.g., \emph{Where should I go when I am thirsty?} $\rightarrow$ water fountain).
  \item \textbf{Spatial Understanding:} Questions requiring spatial reasoning, such as nearest objects or relative positions.
\end{itemize}

\input{fig_inputs/fig_dataset}
\input{tabs/appx_tabs/dataset_statistics}

\textbf{Question Examples.} 
Fig.~\ref{fig:appx_raven_qa_figure_examples} illustrates QA examples from the \emph{Hard} and \emph{Simple} human-ego splits, as well as the \emph{Habitat Simulation} split, in \ourdataset. QA examples from the \emph{Real-World} robot-ego splits are shown in Fig.~\ref{fig:real_robot_rollouts}.

\input{fig_inputs/appx_figs_input/fig_irs_hard_examples}

%% file: fig_inputs/appx_figs_input/fig_three_dataset_rollout.tex
\begin{figure}[ht]
  \centering
  \includegraphics[width=0.9\linewidth]{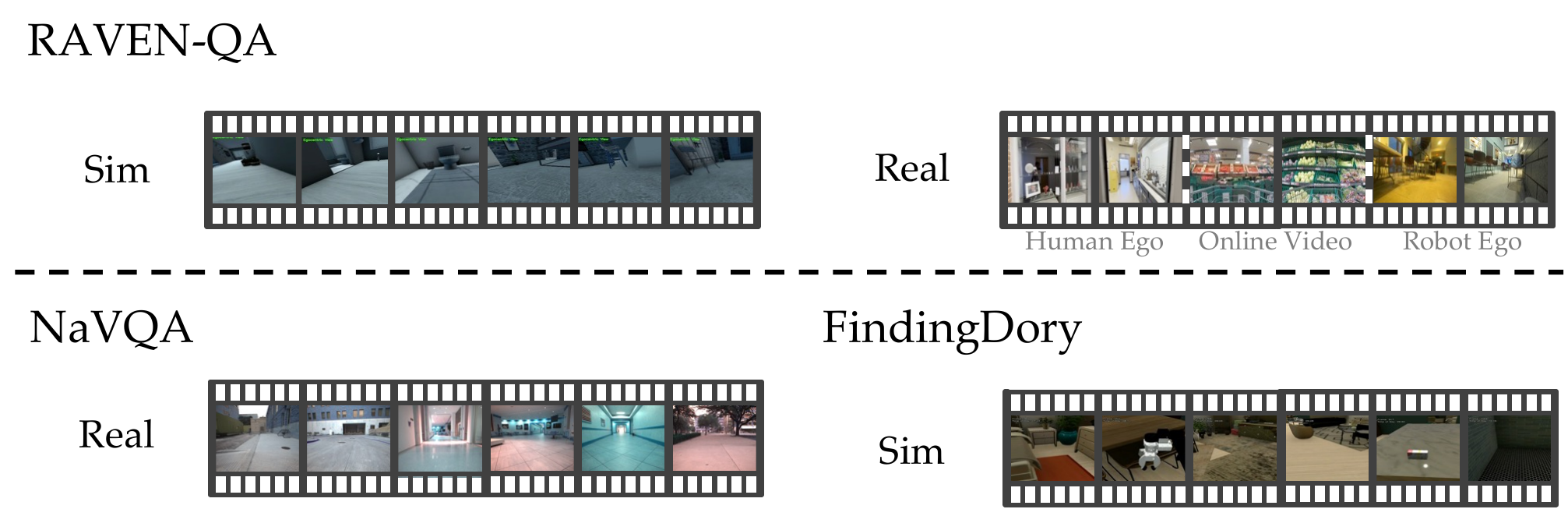}
  \caption{Overview of the evaluation suites. We list the rollout frames in \navqa, \findingdory, and \ourdataset.}
  \label{fig:three_dataset_rollout}
\end{figure}

%% file: fig_inputs/fig_dataset.tex
\begin{figure}[ht]
  \centering
  \includegraphics[width=.8\linewidth]{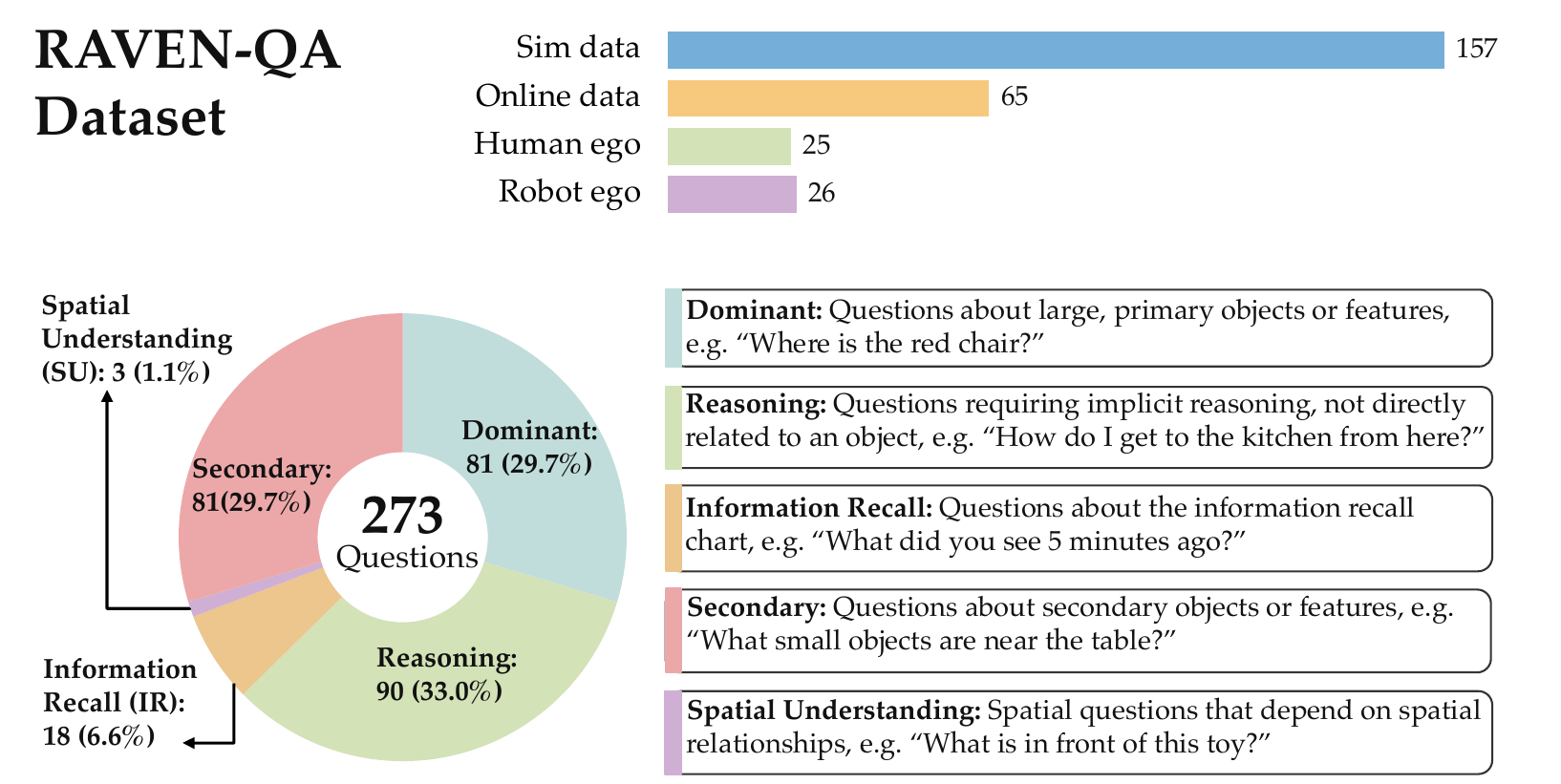}
  \caption{We introduce \ourdataset, which is composed of 273 question and answering pairs. The dataset consists of spatial understanding, reasoning, information recall, dominant and secondary objects retrieval questions.}
  \label{fig:dataset}
\end{figure}

%% file: tabs/appx_tabs/dataset_statistics.tex
\begin{table*}[ht]
\centering
\footnotesize
\caption{RAVEN-QA Statistics across Different Ego Perspectives}
\label{tab:raven_qa_statistics}
\footnotesize
\setlength{\tabcolsep}{3.0pt}
\renewcommand{\arraystretch}{1.08}
\begin{tabular}{lcccccc}
\toprule
& Reasoning & Dominant & Secondary & Information Recall & Spatial Understanding & Total \\
\midrule
Habitat Simulation (Robot Ego)    & 76 & 36 & 45 & 0  & 0 & 157 \\
Web Video (Human Ego) & 5  & 30 & 23 & 7  & 0 & 65  \\
Real-World Exploration (Robot Ego)   & 5  & 9  & 9  & 3  & 0 & 26  \\
Self-Recorded (Human Ego)    & 4  & 6  & 4  & 8  & 3 & 25  \\
\midrule
\textbf{Total}   & \textbf{90} & \textbf{81} & \textbf{81} & \textbf{18} & \textbf{3} & \textbf{273} \\
\bottomrule
\end{tabular}
\end{table*}

%% file: fig_inputs/appx_figs_input/fig_irs_hard_examples.tex
\begin{figure}[ht]
  \centering
  \includegraphics[width=1\linewidth]{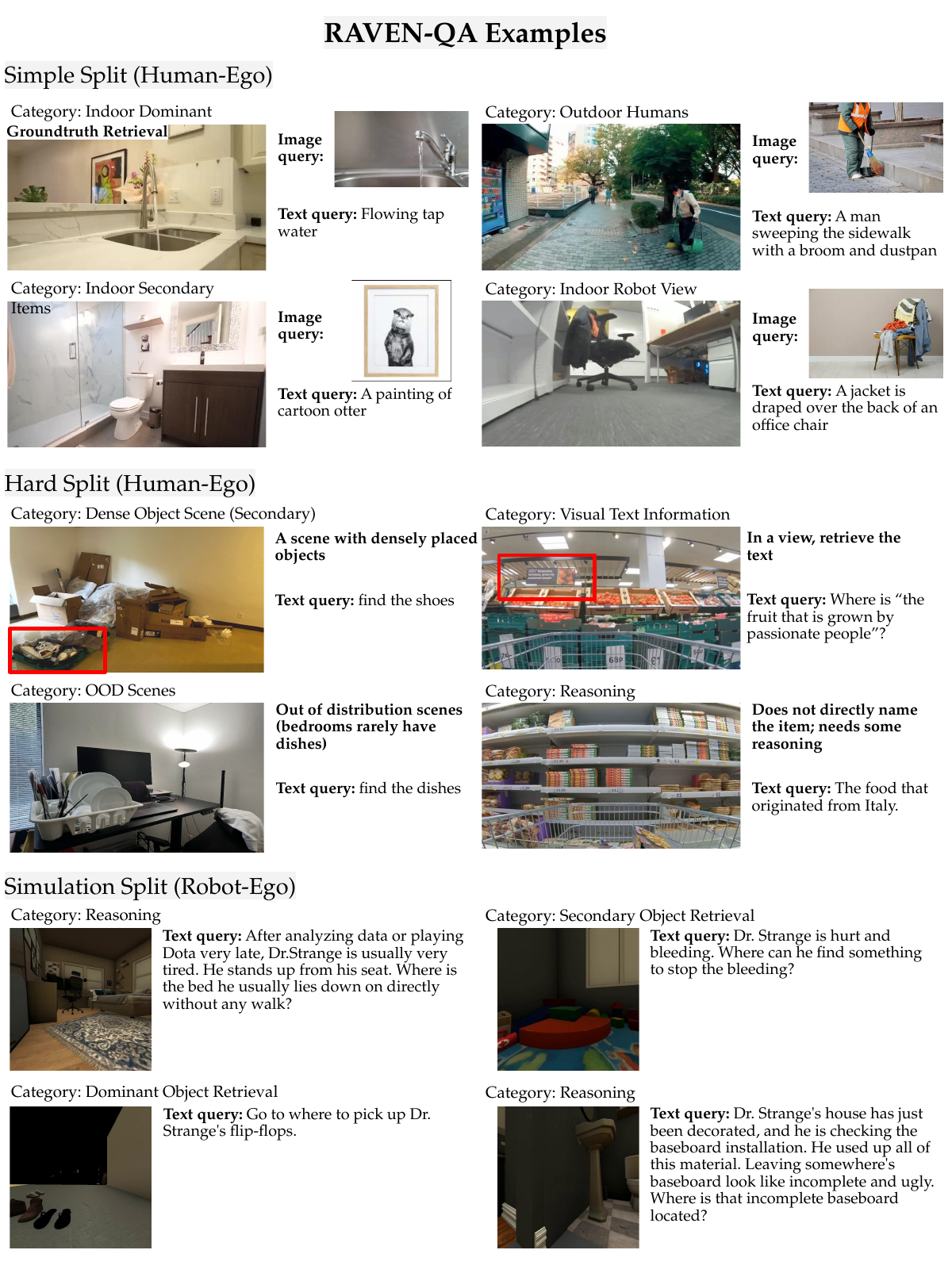}
  \caption{Some various query examples from different splits of \ourdataset, covering different categories, and supporting text-based and image-based queries. For the Real-World Robot split of \ourdataset, please check Fig.~\ref{fig:real_robot_rollouts}.}
  \label{fig:appx_raven_qa_figure_examples}
\end{figure}


%% file: secs/8.B-appendix.tex
\section{Full Experimental Results and Implementation Details}
\label{appx_short:results_and_implement_details}

\vspace{10pt}

\subsection{Full Experimental Results on \navqa}
\label{appx:navqa_full_results}

Full results of \navqa are shown in Table~\ref{tab:exp_navqa_full}. Random method only supports the bianry questions in descriptive question accuracy, therefore, for temporal and positional error, we do not report random method error.

\input{tabs/appx_tabs/exp_navqa_full}
To facilitate quantitative analysis of different model, we convert the positional error and temporal error into accuracy based on this method:
\begin{itemize}
    \item We define a spatial question to be correct if the answer is within 15 meters from ground truth goal.
    \item We define a temporal question to be correct if the answer is within 0.5 mins from ground truth temporal answer.
\end{itemize}
Table~\ref{tab:exp_navqa_overall} shows the results.
\input{tabs/appx_tabs/exp_navqa_overall}

\vspace{10pt}
\subsection{Full Experimental Results on \ourdataset}
\label{appx:sim_full_results}
We present the full, per-category accuracy results of our Habitat Simulation Dataset in Table~\ref{tab:darpa_sim_full}. Fig.~\ref{fig:simulation_example} demonstrates two example scenes from habitat simulation and their 2D occupancy map from exploration phase.
\input{fig_inputs/appx_figs_input/fig_simulation_example}
\input{tabs/darpa_sim_table_full}

To summarize, results tables of \method on \ourdataset, \findingdory, \navqa are shown in Table~\ref{tab:real_and_sim_main_acc_result}, Table~\ref{tab:findingdory_results}, Table~\ref{tab:navqa_results} and Table~\ref{tab:IRS_simple_hard}.
\input{tabs/navqa_table}

\input{tabs/irs_table}

\vspace{10pt}

\subsection{Real-World Implementation Details}

As shown in Fig.~\ref{fig:robot}, our real-world implementation use commercial RGB-D SLAM from ZED 2i camera to provide pointcloud of a 3D scene. And we project it into the 2D occupancy map before use trajectory planner to follow the waypoints to navigate to the frontiers when exploring or to the subgoals when retrieving.
\input{fig_inputs/fig_robot}

As shown in Algorithm~\ref{alg:implementation}, the robot acts in an exploration-execution loop. During the exploration phase, we run real-time RGB-D SLAM with a ZED 2i camera to reconstruct a 3D point cloud of the environment. After being denoised, this 3D point cloud is projected within a fixed height range in \verb|[mix_height, max_height]| onto the ground plane to obtain a 2D occupancy map. In parallel, we maintain an \emph{explored-area} defined by the ZED 2i field of view (FOV) and depth sensing range; \emph{explored-area} is accumulated over time as the robot moves. Using the \emph{explored-area}, we apply a frontier-selection algorithm to identify candidate frontiers along the boundary between explored and unexplored free space. We then select the nearest frontier and plan a collision-free path on a dilated occupancy map using $A^*$, yielding a sequence of waypoints. A low-level PD controller drives the Unitree Go1 to track these waypoints and perform frontier exploration. Full frontier exploration results of four scenes are shown in Fig.~\ref{fig:frontier_exploration}. We show the planned path with waypoints in the public area scenery in Fig.~\ref{fig:waypoints}.
\input{fig_inputs/appx_figs_input/fig_frontier_exploration}
\input{fig_inputs/appx_figs_input/fig_waypoints}

During execution, a target $(x,y)$ goal is determined from \method, \remembr or \vlmonly method. The occupancy map built during exploration is reused and we run $A^*$ to navigate the robot to the target location. Results demonstrating robot following the target goals of Gemini-2.5-Flash Only method are shown in Fig.~\ref{fig:retireval_bowen_vlm_only}.
\input{fig_inputs/appx_figs_input/fig_retireval_bowen_vlm_only}

\begin{algorithm2e}[ht]
\caption{Exploration and Execution for Real-World Navigation}
\label{alg:implementation}
\textbf{Initialization} Query method $\mathcal{M}\in\{\method,\text{\remembr},\text{\vlmonly}\}$\;
\BlankLine
\textbf{Phase I: Frontier-based Exploration}\;
Initialize occupancy map $O \leftarrow \emptyset$ and explored-area $E \leftarrow \emptyset$\;
\While{exploration not terminated}{
    Run RGB-D SLAM to estimate pose $p_t$ and reconstruct point cloud $PC_t$\;
    Capture BGRA image $I_t$ with timestamp $t$\;
    Store $(I_t, p_t, t)$\;
    Denoise $PC_t$ and filter points with height $z \in [h_{\min}, h_{\max}]$\;
    Project filtered points to 2D plane and update occupancy map $O$\;
    Update explored-area $E$ using ZED 2i FOV $\mathcal{F}$ and range $d_{\max}$\;
    Detect candidate frontiers $\mathcal{F}_t \leftarrow \textsc{FrontierSelect}(O, E)$\;
    Choose nearest frontier $f^\star \leftarrow \arg\min_{f\in\mathcal{F}_t}\mathrm{L2\_distance}(f,\mathbf{x}_t)$\;
    Dilate occupancy map $\tilde{O} \leftarrow \textsc{Dilate}(O)$\;
    Plan waypoints list $W \leftarrow \textsc{AStar}(\tilde{O}, \mathbf{x}_t, f^\star)$\;
    Track $W$ \;
}

\BlankLine

\textbf{Question and Answer Design} Extract frames to formulate dataset $Q$\;

\BlankLine



\textbf{Phase II: Query-conditioned Execution}\;
\For{$q_i \in Q$}{
    $(x_i, y_i) \leftarrow \mathcal{M}(q_i)$\;
    Append Predicted goal list $G$ with $(x_i, y_i)$\;
}
\For{$(x_i, y_i) \in G$}{
    Plan a path $W \leftarrow \textsc{AStar}(\tilde{O}, \mathbf{x}, (x,y))$ using the occupancy map\;
    Track $W$ with the PD controller until reaching the goal\;
}
\end{algorithm2e}

\vspace{10pt}
\subsection{Empirical Choice of Parameters}

In our research, we disclose the choice of hyper-parameters, which, we conclude from our empirical experiments, highly depends on the scale of the model--whether it is a small open-source model or a closed-source frontier model. 

\paragraph{Retrieval top-K}
Retrieval top-K refers to the number of memory items returned to the agent in each retrieval call. For the FindingDory benchmark, the context memory scale is up to 3000 frames, therefore we aim to choose greater top-K values for the tested model Gemini-2.5-Flash since it supports a longer visual context. For the RAVEN-QA dataset, after sub-sampling, the memory scale is around 15-30 frames, so to evaluate the efficacy of retrieval, we set a lower value of top-K. Table~\ref{tab:appx_hyperparams_list} shows the assignment of top-K parameters among different benchmarks.\cite{xiao2023efficient}

\paragraph{Maximal number of tool uses}
Too many tool calls greatly increase the context window, making the model difficult to reason about too many items. Therefore, for \ourdataset  and \navqa benchmarks, we follow the original setting in \remembr, which restricts the number of tool calls to 3 at maximum. For \findingdory, we loosen the restriction to 15 since the memory database is much larger. More details are listed in Table~\ref{tab:appx_hyperparams_list}.

\paragraph{Decoding temperature} We did not note significant performance difference for decoding temperature between 0.0001 and 0.7. We used 0.0001, which equals greedy sampling, for web-sourced and self-recorded \ourdataset, and 0.5$\sim$0.7 for simulated and real-world \ourdataset, \navqa, and \findingdory. See Table~\ref{tab:appx_hyperparams_list} and for details.

\input{tabs/appx_tabs/params}

\vspace{10pt}
\subsection{Exploring Image-Query Retrieval with Multimodal Embeddings}
While the main paper focused on text-based queries, this section introduces images as a distinct query modality. We demonstrate that by leveraging multimodal embeddings, image-based retrieval achieves performance comparable to that of standard text queries.

\paragraph{Text- and Image-query Image Reverse Search (T-IRS \& I-IRS)}
We investigate multimodal embedder performance in a retrieval framework using text and image queries against subsampled video sequences (Fig.~\ref{fig:appx_pilot_study}). The model maps both sequences and queries into a unified latent space, where FAISS \cite{douze2024faiss} or Milvus \cite{10.1145/3448016.3457550} identifies the nearest neighbors via cosine distance. Our benchmark includes five curated tour videos and one robotic-view sequence, encompassing 54 queries across dominant object, secondary object, and outdoor human retrieval tasks (Examples are in Fig.~\ref{fig:appx_raven_qa_figure_examples}).

\input{fig_inputs/appx_figs_input/fig_pilot_study}
\input{fig_inputs/appx_figs_input/fig_irs_query_examples}

\paragraph{Embedding Model Candidates}
We evaluate candidate embedders across four training paradigms: 
(i) \textit{contrastive}: CLIP \cite{radford2021learningtransferablevisualmodels}, SigLIP \cite{zhai2023sigmoidlosslanguageimage}, and QQMM-Embed-v2 \cite{xue2025improvemultimodalembeddinglearning}; 
(ii) \textit{autoregressive}: PaliGemma2 \cite{steiner2024paligemma2familyversatile} and Qwen2.5-VL \cite{Qwen2.5-VL}; 
(iii) \textit{self-distillation}: Dinov3 \cite{simeoni2025dinov3}; and 
(iv) \textit{closed-source}: Seed-1.6-Embedding \cite{seed16embedding} and Google Multimodal Embedding. 
We also include a random retriever as a performance baseline.

\paragraph{Results}
T-IRS and I-IRS performance in Table~\ref{tab:merged_pilot_accuracy_results} reveals that SOTA embedders like QQMM-v2 achieve image-query performance parity with text-based retrieval. To analyze retrieval quality, Table~\ref{tab:appx_embedding_closer_look} reports the average similarity ratio ($S_{top1}/S_{top2}$) for successes and the mean rank of ground-truth items for Top-1 failures. We find that robust embeddings: (i) maintain larger similarity margins during successful retrievals, and (ii) keep ground-truth candidates near the top of the rank during failures. These metrics validate our reliance on QQMM and Seed architectures for subsequent experiments. 

\input{tabs/appx_tabs/pilot_study_table_text_q}
\input{tabs/appx_tabs/pilot_study_table_image_q}
\input{tabs/appx_tabs/pilot_study_top_k_closer_look}







\paragraph{Key Takeaways} 
We summarize our findings as follows:

\quad\quad\quad(i) Open-source embeddings have achieved performance parity with closed-source models.

\quad\quad\quad(ii) Robust embeddings successfully capture non-dominant objects within complex scenes.

\quad\quad\quad(iii) Raw VLM hidden states are poorly suited for retrieval compared to specialized embedding recipes.

\quad\quad\quad(iv) QQMM-v2 and Seed-1.6 yield the highest accuracy for both image and text queries.

\quad\quad\quad(v) SigLIP provides the best performance-to-efficiency ratio for on-device applications.

\vspace{10pt}
\subsection{More Justifications on \method Framework Design}
We design the \method framework based on several key considerations. In this subsection, we elaborate on the underlying insights that motivate our design choices.

\paragraph{Performance improves with appropriate top-$K$}

We conducted preliminary experiments on the \emph{Simple} split of \ourdataset, which consists of 54 questions. The experiments were performed using a frontier VLM (Gemini-2.5-Flash) together with a strong multimodal embedding model (Seed1.6-Embed). The results are shown in Fig.~\ref{fig:appx_topk_simi_score}. For frontier closed-source models, retrieving more memory items—i.e., using a larger top-$K$—consistently improves agent performance. Based on these observations, we set $K=5$ for \ourdataset in our benchmark and select different $K$ values for other datasets according to their memory scale.

\input{fig_inputs/appx_figs_input/framework_study_topk_sim_scores}

\paragraph{Providing similarity scores improves VLM reasoning}

When composing the retrieval response for the VLM agent, it is non-trivial to decide what information should be exposed. In \remembr{}, similarity scores are not provided to the language model. Based on our empirical observations, we include similarity scores alongside each retrieved memory item in the response. As shown in Fig.~\ref{fig:appx_topk_simi_score}, this additional signal consistently benefits the agent’s subsequent reasoning. We conjecture that similarity scores offer a useful confidence cue, enabling the VLM to better assess the relevance of retrieved memories. This observation provides practical guidance for optimizing the information composition of retrieval responses.

\paragraph{Chronological ordering of the retrieved memories is more effective}

For text- and position-based queries, retrieval results are ranked by similarity, with the most relevant memories listed first. However, we observe that for time-based retrieval, especially under large memory scales and larger top-$k$, an alternative ordering strategy enables more effective interpretation by the agent. While distance-based ranking is the default in the original \remembr{} implementation~\cite{anwar2024remembrbuildingreasoninglonghorizon} and in many RAG systems, its rationale is rarely examined. We argue that a more suitable strategy is to align retrieval ordering with the model’s training paradigm. Modern multimodal models are trained on massive video corpora and are inherently optimized to model causal and temporal dynamics. As a result, chronological structure provides a more natural inductive bias than similarity-based ordering for time-centric queries. Accordingly, when the agent issues a time-based retrieval request, we return a temporally ordered list of memories starting from the target timestamp, rather than ranking them purely by distance. We do not center the target timestamp in the returned sequence, as doing so may cause memories near the requested time to receive less attention due to the attention sink issue~\cite{xiao2023efficient}. The intuition behind this design is illustrated in Fig.~\ref{fig:appx_retrieval_ordering_intuition} and is supported by the empirical results in Table~\ref{tab:appx_ordering_exp_results} on \findingdory.

\input{fig_inputs/appx_figs_input/fig_ordering_intuition}

\input{tabs/appx_tabs/ordering_exp_results}

\subsection{More detailed Analyses of \method}
\label{appx:more_detailed_analyses}
\textbf{\method exhibits great compactness of memory} We conducted analysis of the compressibility of RAVEN memory in a real-world retrieval experiment. Fig.~\ref{fig:compression} summarizes the result: owing to powerful visual embeddings that can capture fine-grained details, RAVEN is able to maintain $90+\%$ performance while undergoing $250\times$ sparsification. When compared to a raw video stream, RAVEN takes $22,315\times$ less memory, while still preserving necessary information for navigation memory.
\input{fig_inputs/rebuttal_fig_inputs/judgement_accuracy_vs_stride}

\textbf{\method exhibits greater consistency across repeated runs.} As shown in Table~\ref{tab:real_and_sim_main_acc_result}, \vlmonly suffers from noticeably higher variance ($\pm1\sim6\%$), which can undermine reliability in real-world deployments. In comparison, \method consistently achieves stable performance across different sampling seeds ($0\sim5\%$). This robustness arises from its structured, multi-step agentic reasoning with iterative retrieval and verification, whereas \vlmonly relies on a single-pass generation that is inherently less stable and less trustworthy.

\textbf{Robustness to exploration completeness} We study the performance of RAVEN with incomplete exploration, which is likely in complex environments. Fig.~\ref{fig:rebuttal_coverage_experiment} shows that even at \(\sim\)20\% footprint coverage, \method can achieve near-perfect task accuracy. This is because \method operates on visual observations and embeddings and does not require full state visitation to register objects.
\input{fig_inputs/rebuttal_fig_inputs/rebuttal_coverage_experiment}

%% file: tabs/appx_tabs/exp_navqa_full.tex
\begin{center}
\fontsize{6.8}{7.6}\selectfont
\setlength{\tabcolsep}{3pt}
\renewcommand{\arraystretch}{1.1}

\begin{longtable}{llccc ccc ccc}
\caption{Full NaVQA results across different VLMs including Gemini-2.5-Flash, Gemini-2.5-Pro, Gemini-3-Pro, GPT-4o, GPT-5, GPT-5.2 on three methods: \method, \remembr and VLM only. We report descriptive question accuracy (\%), positional error (m), and temporal error (min) on Short (S), Medium (M), and Long (L) horizons as Mean$\pm$Std. QQMM was evaluated using 4-bit quantization due to GPU memory constraints on the RTX 5070 Ti.}
\label{tab:exp_navqa_full} \\

\toprule
\textbf{LLM} & \textbf{Method / Encoder} &
\multicolumn{3}{c}{\textbf{Descriptive Accuracy (\%)}} &
\multicolumn{3}{c}{\textbf{Positional Error (m)}} &
\multicolumn{3}{c}{\textbf{Temporal Error (min)}} \\
\cmidrule(lr){3-5}\cmidrule(lr){6-8}\cmidrule(lr){9-11}
& & S & M & L & S & M & L & S & M & L \\
\midrule
\endfirsthead

\toprule
\textbf{LLM} & \textbf{Method / Encoder} &
S & M & L & S & M & L & S & M & L \\
\midrule
\endhead

\midrule
\multicolumn{11}{r}{\textit{Continued on next page}} \\
\endfoot

\bottomrule
\endlastfoot

& \multicolumn{1}{l}{\textbf{\method}} \\
\multirow{10}{*}{\textbf{Gemini-2.5-Flash}} & Seed1.6-Embed
& 41.70 & 42.40 & 50.00
& 25.31$\pm$35.96 & 60.51$\pm$86.97 & 117.60$\pm$125.93
& 0.33$\pm$0.30 & 1.40$\pm$2.00 & 3.60$\pm$5.58 \\

& QQMM-v2
& 51.20 & 47.40 & 60.00
& \textbf{4.49$\pm$8.53} & 16.20$\pm$22.50 & 49.51$\pm$61.24
& 0.21$\pm$0.28 & \textbf{0.67$\pm$1.63} & 1.55$\pm$2.74 \\

& SigLIP
& \textbf{63.40} & 47.40 & 60.00
& 7.39$\pm$8.40 & 26.59$\pm$36.45 & 47.53$\pm$64.75
& 0.24$\pm$0.33 & 1.04$\pm$1.77 & 0.72$\pm$1.21 \\

& CLIP-B
& 51.20 & 50.00 & 55.00
& 9.84$\pm$10.27 & 23.62$\pm$31.82 & 58.07$\pm$58.11
& 0.21$\pm$0.21 & 1.01$\pm$1.68 & 2.48$\pm$5.46 \\

& \multicolumn{1}{l}{\textbf{VLM Only}} \\
& ----
& 58.50 & 50.00 & 60.00
& 7.78$\pm$7.60 & \textbf{12.48$\pm$21.20} & 40.82$\pm$52.72
& \textbf{0.19$\pm$0.18} & 1.19$\pm$2.06 & 1.10$\pm$2.88 \\

& \multicolumn{1}{l}{\textbf{\remembr}} \\

& MixedBread 
& 59.50 & \textbf{60.50} & \textbf{75.00}
& 9.10$\pm$10.84 & 28.89$\pm$36.21 & \textbf{40.72$\pm$53.78}
& 0.20$\pm$0.26 & 2.01$\pm$2.07 & \textbf{0.62$\pm$1.19} \\

& Seed1.6-Embed
& 26.20 & 26.30 & 25.00
& 21.29$\pm$34.10 & 19.69$\pm$25.34 & 56.27$\pm$76.91
& 0.25$\pm$0.19 & 1.47$\pm$1.83 & 2.34$\pm$4.89 \\

& QQMM-v2 
& 19.00 & 18.40 & 5.00
& 9.46$\pm$13.62 & 26.01$\pm$27.93 & 43.01$\pm$53.63
& 0.43$\pm$0.29 & 0.94$\pm$1.63 & 1.62$\pm$4.66 \\

& CLIP-B 
& 28.60 & 34.20 & 15.00
& 19.80$\pm$14.27 & 51.66$\pm$36.47 & 67.26$\pm$69.98
& 0.50$\pm$0.25 & 1.38$\pm$1.66 & 2.17$\pm$3.48 \\

& SigLIP 
& 31.00 & 31.60 & 25.00
& 17.45$\pm$15.72 & 37.77$\pm$31.81 & 65.93$\pm$70.48
& 0.59$\pm$0.40 & 1.25$\pm$1.08 & 3.37$\pm$4.97 \\

\cmidrule(lr){1-2}
& \multicolumn{1}{l}{\textbf{\method}} \\
\multirow{10}{*}{\textbf{Gemini-2.5-Pro}} & Seed1.6-Embed
& 62.90 & 57.60 & 55.60
& 15.37$\pm$12.17 & 57.04$\pm$79.61 & 93.96$\pm$105.42
& 0.34$\pm$0.28 & 1.72$\pm$2.16 & 4.33$\pm$5.60 \\

& QQMM-v2
& \textbf{69.00} & 60.50 & \textbf{75.00}
& \textbf{3.79$\pm$6.19} & \textbf{12.29$\pm$21.01} & 44.02$\pm$57.13
& 0.29$\pm$0.34 & \textbf{0.69$\pm$1.71} & 0.88$\pm$1.73 \\

& SigLIP
& 66.70 & 57.90 & 70.00
& 7.06$\pm$9.22 & 20.08$\pm$27.32 & 37.53$\pm$49.41
& 0.24$\pm$0.31 & 0.86$\pm$1.62 & 1.68$\pm$4.40 \\

& CLIP-B
& 61.90 & 55.30 & 45.00
& 12.35$\pm$12.42 & 22.62$\pm$33.82 & 43.24$\pm$43.05
& 0.24$\pm$0.31 & 0.86$\pm$1.62 & 1.68$\pm$4.40 \\

& \multicolumn{1}{l}{\textbf{VLM Only}} \\
& ----
& 66.70 & 60.50 & 65.00
& 3.90$\pm$4.04 & 12.35$\pm$20.81 & \textbf{24.93$\pm$47.88}
& 0.11$\pm$0.08 & 0.98$\pm$2.07 & \textbf{0.16$\pm$0.12} \\

& \multicolumn{1}{l}{\textbf{\remembr}} \\
& MixedBread
& 66.70 & \textbf{68.40} & 70.00
& 6.11$\pm$8.42 & 28.25$\pm$36.19 & 36.71$\pm$52.83
& 0.16$\pm$0.26 & 1.38$\pm$1.89 & 1.85$\pm$4.62 \\

& Seed1.6-Embed
& 19.00 & 28.90 & 20.00
& 8.20$\pm$9.04 & 23.57$\pm$28.32 & 36.44$\pm$49.86
& 0.24$\pm$0.25 & 1.20$\pm$1.80 & 2.39$\pm$5.11 \\

& QQMM-v2
& 21.40 & 26.30 & 5.00
& 9.43$\pm$11.32 & 31.14$\pm$37.17 & 31.56$\pm$44.88
& \textbf{0.08$\pm$0.07} & 1.04$\pm$1.83 & 2.05$\pm$4.71 \\

& CLIP-B
& 26.20 & 34.20 & 0.00
& 16.08$\pm$13.48 & 48.19$\pm$34.59 & 74.10$\pm$83.33
& 0.29$\pm$0.24 & 1.25$\pm$1.39 & 2.43$\pm$3.41 \\

& SigLIP
& 31.00 & 34.20 & 10.00
& 16.54$\pm$15.46 & 45.12$\pm$30.67 & 61.98$\pm$53.82
& 0.39$\pm$0.21 & 1.38$\pm$1.49 & 3.44$\pm$5.08 \\

\cmidrule(lr){1-2}
& \multicolumn{1}{l}{\textbf{\method}} \\
\multirow{10}{*}{\textbf{Gemini-3-Pro}}& Seed1.6-Embed
& \textbf{76.20} & 52.60 & 75.00
& 4.58$\pm$6.08 & 17.74$\pm$26.58 & 42.63$\pm$60.47
& 0.20$\pm$0.29 & 0.78$\pm$1.74 & 0.43$\pm$1.51 \\

& QQMM-v2
& \textbf{76.20} & 52.60 & \textbf{80.00}
& \textbf{2.37$\pm$3.89} & \textbf{11.48$\pm$20.88} & 38.81$\pm$56.57
& 0.23$\pm$0.31 & 0.88$\pm$1.75 & \textbf{0.07$\pm$0.09} \\

& SigLIP
& 61.90 & 57.90 & \textbf{80.00}
& 6.51$\pm$9.15 & 14.09$\pm$19.62 & 33.20$\pm$49.23
& 0.16$\pm$0.16 & 0.81$\pm$1.76 & 1.17$\pm$4.62 \\

& CLIP-B
& 66.70 & 57.90 & 70.00
& 7.15$\pm$10.89 & 16.21$\pm$23.27 & 43.86$\pm$51.62
& 0.09$\pm$0.10 & 0.79$\pm$1.75 & 1.19$\pm$3.71 \\

& \multicolumn{1}{l}{\textbf{VLM Only}} \\
& ----
& 66.70 & 63.20 & \textbf{80.00}
& 4.81$\pm$6.69 & 11.87$\pm$20.89 & \textbf{27.10$\pm$49.71}
& 0.21$\pm$0.31 & 0.73$\pm$1.71 & 0.15$\pm$0.28 \\

& \multicolumn{1}{l}{\textbf{\remembr}} \\
& MixedBread 
& 66.70 & \textbf{73.70} & 75.00
& 2.78$\pm$2.94 & 24.52$\pm$36.14 & 43.00$\pm$55.50
& \textbf{0.04$\pm$0.03} & 1.08$\pm$2.05 & 1.22$\pm$4.52 \\

& Seed1.6-Embed
& 59.50 & 60.50 & 60.00
& 8.36$\pm$11.15 & 18.09$\pm$25.59 & 37.04$\pm$56.94
& 0.24$\pm$0.26 & 0.81$\pm$1.58 & 1.61$\pm$4.71 \\

& QQMM-v2 
& 57.10 & 60.50 & 65.00
& 7.64$\pm$10.64 & 22.04$\pm$29.49 & 35.41$\pm$51.04
& 0.09$\pm$0.10 & \textbf{0.67$\pm$1.22} & 1.81$\pm$4.70 \\

& CLIP-B 
& 57.10 & 60.50 & 70.00
& 16.69$\pm$14.35 & 38.29$\pm$33.48 & 74.30$\pm$78.75
& 0.36$\pm$0.20 & 1.09$\pm$1.46 & 3.47$\pm$5.36 \\

& SigLIP 
& 59.50 & 63.20 & 70.00
& 16.40$\pm$16.57 & 37.25$\pm$28.25 & 59.02$\pm$61.46
& 0.47$\pm$0.37 & 0.99$\pm$1.47 & 2.78$\pm$4.77 \\

\cmidrule(lr){1-2}
& \multicolumn{1}{l}{\textbf{\method}} \\
\multirow{10}{*}{\textbf{GPT-4o}}& Seed1.6-Embed
& \textbf{64.70} & 42.40 & 61.10
& 4.49$\pm$5.35 & 25.33$\pm$29.93 & 53.63$\pm$63.61
& 0.21$\pm$0.16 & 2.35$\pm$1.98 & 1.81$\pm$2.97 \\

& QQMM-v2
& 62.50 & 47.40 & \textbf{70.00}
& \textbf{3.99$\pm$6.19} & 17.15$\pm$22.98 & \textbf{38.63$\pm$54.09}
& 0.19$\pm$0.21 & 1.23$\pm$2.05 & 2.05$\pm$4.74 \\

& SigLIP
& 60.60 & 54.50 & 61.10
& 4.24$\pm$5.52 & 19.88$\pm$27.41 & 60.01$\pm$80.14
& 0.29$\pm$0.39 & 1.30$\pm$1.90 & 2.22$\pm$2.96 \\

& CLIP-B
& 60.60 & 48.50 & 55.60
& 8.83$\pm$8.90 & 17.71$\pm$21.47 & 50.86$\pm$46.47
& \textbf{0.16$\pm$0.14} & 1.90$\pm$1.85 & 2.00$\pm$2.88 \\

& \multicolumn{1}{l}{\textbf{VLM Only}} \\
& ----
& 63.40 & 60.50 & 65.00
& 7.40$\pm$10.99 & \textbf{12.56$\pm$17.19} & 40.33$\pm$47.89
& 17.61$\pm$32.61 & \textbf{1.00$\pm$2.06} & 2.68$\pm$6.82 \\

& \multicolumn{1}{l}{\textbf{\remembr}} \\
& MixedBread
& 57.10 & \textbf{63.20} & 65.00
& 6.10$\pm$8.85 & 29.25$\pm$38.34 & 47.22$\pm$59.33
& 0.29$\pm$0.40 & 1.22$\pm$1.78 & \textbf{1.27$\pm$2.16} \\

& Seed1.6-Embed
& 16.70 & 23.70 & 10.00
& 33.39$\pm$83.06 & 30.04$\pm$32.49 & 71.13$\pm$85.31
& 0.40$\pm$0.37 & 1.82$\pm$2.04 & 3.32$\pm$6.45 \\

& QQMM-v2
& 19.00 & 23.70 & 10.00
& 25.06$\pm$35.95 & 42.69$\pm$54.92 & 63.43$\pm$76.03
& 1.45$\pm$3.49 & 2.17$\pm$1.97 & 4.74$\pm$6.87 \\

& CLIP-B
& 14.30 & 26.30 & 5.00
& 33.23$\pm$53.94 & 62.79$\pm$38.99 & 93.55$\pm$76.11
& 1.57$\pm$3.18 & 2.06$\pm$1.76 & 5.25$\pm$5.59 \\

& SigLIP
& 14.30 & 23.70 & 5.00
& 22.60$\pm$25.49 & 53.24$\pm$35.44 & 85.45$\pm$77.89
& 0.35$\pm$0.18 & 1.58$\pm$1.82 & 6.04$\pm$5.87 \\

\cmidrule(lr){1-2}
& \multicolumn{1}{l}{\textbf{\method}} \\
\multirow{10}{*}{\textbf{GPT-5}} & Seed1.6-Embed
& \textbf{80.60} & 63.60 & 66.70
& 6.58$\pm$11.38 & 28.37$\pm$36.75 & 45.56$\pm$58.39
& \textbf{0.08$\pm$0.05} & 1.70$\pm$2.22 & 1.49$\pm$2.45 \\

& QQMM-v2
& 71.40 & 60.50 & 75.00
& \textbf{3.44$\pm$5.28} & \textbf{16.27$\pm$27.37} & \textbf{34.29$\pm$55.70}
& 0.10$\pm$0.10 & 1.25$\pm$2.04 & \textbf{0.80$\pm$1.25} \\

& SigLIP
& 72.20 & 66.70 & 83.30
& 6.37$\pm$8.14 & 21.90$\pm$29.13 & 39.82$\pm$47.95
& 0.18$\pm$0.29 & 1.73$\pm$1.91 & 2.97$\pm$5.23 \\

& CLIP-B
& 75.00 & 63.60 & 61.10
& 7.23$\pm$8.30 & 23.29$\pm$30.62 & 43.74$\pm$47.69
& 0.22$\pm$0.32 & 1.50$\pm$2.00 & 0.96$\pm$2.18 \\

& \multicolumn{1}{l}{\textbf{VLM Only}} \\

& ----
& 73.80 & 60.50 & \textbf{85.00}
& 4.31$\pm$4.66 & 16.56$\pm$28.37 & 37.06$\pm$52.94
& 0.09$\pm$0.06 & \textbf{1.08$\pm$2.06} & 1.46$\pm$4.67 \\

& \multicolumn{1}{l}{\textbf{\remembr}} \\
& MixedBread
& 69.00 & \textbf{68.40} & 80.00
& 7.95$\pm$12.98 & 27.88$\pm$40.25 & 38.75$\pm$57.64
& 0.18$\pm$0.26 & 1.14$\pm$2.02 & 1.21$\pm$2.14 \\

& Seed1.6-Embed
& 23.80 & 23.70 & 15.00
& 13.47$\pm$13.70 & 31.29$\pm$45.26 & 47.24$\pm$61.24
& 0.50$\pm$0.23 & 1.38$\pm$1.63 & 3.88$\pm$6.40 \\

& QQMM-v2 
& 26.20 & 26.30 & 15.00
& 10.91$\pm$13.90 & 42.19$\pm$47.31 & 54.13$\pm$64.48
& 0.36$\pm$0.29 & 1.30$\pm$1.58 & 3.55$\pm$6.08 \\

& CLIP-B 
& 26.20 & 26.30 & 20.00
& 16.78$\pm$13.67 & 50.49$\pm$45.25 & 55.73$\pm$62.35
& 0.46$\pm$0.21 & 1.74$\pm$1.84 & 4.31$\pm$5.26 \\

& SigLIP 
& 28.60 & 23.70 & 10.00
& 18.83$\pm$15.24 & 47.45$\pm$49.96 & 69.26$\pm$73.53
& 0.49$\pm$0.27 & 1.20$\pm$1.59 & 4.22$\pm$5.27 \\

\cmidrule(lr){1-2}
& \multicolumn{1}{l}{\textbf{\method}} \\
\multirow{10}{*}{\textbf{GPT-5.2}}& Seed1.6-Embed
& 65.90 & 50.00 & 75.00
& 6.18$\pm$8.33 & 28.62$\pm$35.97 & 51.61$\pm$66.33
& 0.26$\pm$0.29 & 1.31$\pm$2.02 & 1.53$\pm$2.61 \\

& QQMM-v2
& 68.30 & 52.60 & 75.00
& \textbf{2.93$\pm$5.53} & 19.32$\pm$27.25 & 45.20$\pm$59.78
& 0.13$\pm$0.17 & 1.20$\pm$2.06 & 2.63$\pm$5.20 \\

& SigLIP
& 63.40 & 55.30 & 75.00
& 6.98$\pm$9.22 & 22.58$\pm$29.55 & 38.12$\pm$47.82
& 0.23$\pm$0.28 & 1.33$\pm$2.05 & 3.83$\pm$6.36 \\

& CLIP-B
& 53.70 & 52.60 & 65.00
& 11.68$\pm$12.96 & 23.15$\pm$28.54 & 51.03$\pm$49.81
& 0.30$\pm$0.31 & 1.12$\pm$1.96 & 4.19$\pm$6.90 \\

& \multicolumn{1}{l}{\textbf{VLM Only}} \\
& ----
& 56.10 & 52.60 & \textbf{85.00}
& 5.45$\pm$9.02 & \textbf{16.68$\pm$21.84} & \textbf{34.59$\pm$42.09}
& \textbf{0.10$\pm$0.09} & \textbf{1.08$\pm$2.06} & \textbf{0.56$\pm$1.23} \\

& \multicolumn{1}{l}{\textbf{\remembr}} \\
& MixedBread
& \textbf{69.00} & \textbf{68.40} & 75.00
& 6.27$\pm$7.33 & 29.10$\pm$40.03 & 38.37$\pm$50.84
& 0.25$\pm$0.36 & 1.77$\pm$2.12 & 2.58$\pm$5.18 \\

& Seed1.6-Embed
& 14.30 & 23.70 & 15.00
& 11.57$\pm$11.21 & 43.09$\pm$50.87 & 63.16$\pm$64.69
& 0.46$\pm$0.34 & 1.74$\pm$1.87 & 4.62$\pm$7.09 \\

& QQMM-v2
& 21.40 & 28.90 & 10.00
& 18.39$\pm$16.36 & 26.77$\pm$28.26 & 68.32$\pm$74.63
& 0.45$\pm$0.34 & 1.91$\pm$2.19 & 3.85$\pm$6.53 \\

& CLIP-B
& 19.20 & 22.20 & 7.10
& 19.12$\pm$13.37 & 53.31$\pm$38.54 & 80.63$\pm$70.35
& 0.59$\pm$0.32 & 1.81$\pm$2.09 & 5.56$\pm$5.59 \\

& SigLIP
& 16.70 & 34.20 & 5.00
& 22.14$\pm$16.27 & 40.68$\pm$35.62 & 74.20$\pm$68.03
& 0.48$\pm$0.21 & 1.82$\pm$1.91 & 4.96$\pm$5.27 \\

\cmidrule(lr){1-2}
\multicolumn{2}{c}{\textbf{Random}}
& 30.95 & 35.53 & 35.00
& ---- & ---- & ----
& ---- & ---- & ----
\end{longtable}
\end{center}

%% file: tabs/appx_tabs/exp_navqa_overall.tex

{\scriptsize
\begin{longtable}{llcccc}
\caption{Summary accuracies (\%) (Overall / Descriptive / Positional / Temporal) grouped by VLM.}%
\label{tab:exp_navqa_overall}\\
\toprule
\textbf{VLM} & \textbf{Method / Encoder} & \textbf{Overall Accuracy} &
\textbf{Descriptive Accuracy} & \textbf{Positional Accuracy} & \textbf{Temporal Accuracy} \\
\midrule
\endfirsthead

\toprule
\textbf{VLM} & \textbf{Method / Encoder} & \textbf{Overall Accuracy} & \textbf{Descriptive Accuracy} & \textbf{Positional Accuracy} & \textbf{Temporal Accuracy} \\
\midrule
\endhead

\midrule
\multicolumn{6}{r}{\textit{Continued on next page}} \\
\endfoot

\bottomrule
\endlastfoot

& \multicolumn{1}{l}{\textbf{\remembr}} \\
\multirow{10}{*}{\textbf{Gemini-2.5-Flash}}
 & CLIP & 32.9 & 28.0 & 29.6 & 51.3 \\
 & QQMM-v2 & 41.9 & 16.0 & 62.0 & 71.8 \\
 & Seed1.6-Embed & 41.9 & 26.0 & 54.9 & 59.0 \\
 & SigLIP & 32.4 & 30.0 & 33.8 & 35.9 \\
 & mxbai & 61.9 & \textbf{63.0} & 57.7 & 66.7 \\
& \multicolumn{1}{l}{\textbf{\method}} \\
 & CLIP & 54.5 & 51.5 & 50.7 & 69.2 \\
 & QQMM-v2 & 61.7 & 51.5 & \textbf{66.2} & 79.5 \\
 & Seed1.6-Embed & 44.4 & 43.7 & 35.6 & 61.8 \\
 & SigLIP & 63.2 & 56.6 & \textbf{66.2} & 74.4 \\
 & \textbf{VLM Only} & \textbf{63.6} & 55.6 & 64.8 & \textbf{82.1} \\

\cmidrule(lr){1-2}
& \multicolumn{1}{l}{\textbf{\remembr}} \\
\multirow{10}{*}{\textbf{Gemini-2.5-Pro}}
 & CLIP & 31.9 & 24.0 & 33.8 & 48.7 \\
 & QQMM-v2 & 43.8 & 20.0 & 59.2 & 76.9 \\
 & Seed1.6-Embed & 43.8 & 23.0 & 62.0 & 64.1 \\
 & SigLIP & 31.0 & 28.0 & 28.2 & 43.6 \\
 & mxbai & 67.6 & \textbf{68.0} & 64.8 & 71.8 \\
& \multicolumn{1}{l}{\textbf{\method}} \\
 & CLIP & 58.4 & 56.0 & 52.1 & 76.3 \\
 & QQMM-v2 & 70.5 & 67.0 & 70.4 & 79.5 \\
 & Seed1.6-Embed & 50.8 & 59.3 & 40.7 & 47.1 \\
 & SigLIP & 67.6 & 64.0 & 67.6 & 76.9 \\
 & \textbf{VLM Only} & \textbf{74.8} & 64.0 & \textbf{80.3} & \textbf{92.3} \\

\cmidrule(lr){1-2}
& \multicolumn{1}{l}{\textbf{\remembr}} \\
\multirow{10}{*}{\textbf{Gemini-3-Pro-preview}}
 & CLIP & 50.5 & 61.0 & 33.8 & 53.8 \\
 & QQMM-v2 & 66.2 & 60.0 & 66.2 & 82.1 \\
 & Seed1.6-Embed & 65.7 & 60.0 & 67.6 & 76.9 \\
 & SigLIP & 50.0 & 63.0 & 33.8 & 46.2 \\
 & mxbai & 74.3 & \textbf{71.0} & 71.8 & 87.2 \\
& \multicolumn{1}{l}{\textbf{\method}} \\
 & CLIP & 66.7 & 64.0 & 60.6 & 84.6 \\
 & QQMM-v2 & \textbf{75.2} & 68.0 & \textbf{78.9} & 87.2 \\
 & Seed1.6-Embed & 72.4 & 67.0 & 71.8 & 87.2 \\
 & SigLIP & 71.4 & 64.0 & 71.8 & \textbf{89.7} \\
 & \textbf{VLM Only} & \textbf{75.2} & 68.0 & \textbf{78.9} & 87.2 \\

\cmidrule(lr){1-2}
& \multicolumn{1}{l}{\textbf{\remembr}} \\
\multirow{10}{*}{\textbf{GPT-4o}}
 & CLIP & 18.4 & 14.5 & 18.3 & 25.6 \\
 & QQMM-v2 & 32.9 & 19.0 & 43.7 & 48.7 \\
 & Seed1.6-Embed & 34.8 & 18.0 & 47.9 & 53.8 \\
 & SigLIP & 23.8 & 16.0 & 29.6 & 33.3 \\
 & mxbai & 61.9 & \textbf{61.0} & 63.4 & 61.5 \\
& \multicolumn{1}{l}{\textbf{\method}} \\
 & CLIP & 50.8 & 54.8 & 55.9 & 32.4 \\
 & QQMM-v2 & \textbf{64.4} & 58.2 & \textbf{67.6} & 74.4 \\
 & Seed1.6-Embed & 55.1 & 55.3 & 62.7 & 41.2 \\
 & SigLIP & 53.1 & 58.3 & 61.0 & 26.5 \\
 & \textbf{VLM Only} & 41.8 & 12.1 & 64.8 & \textbf{76.3} \\

\cmidrule(lr){1-2}
& \multicolumn{1}{l}{\textbf{\remembr}} \\
\multirow{10}{*}{\textbf{GPT-5}}
 & CLIP & 32.4 & 25.0 & 36.6 & 43.6 \\
 & QQMM-v2 & 38.6 & 24.0 & 49.3 & 56.4 \\
 & Seed1.6-Embed & 38.6 & 22.0 & 53.5 & 53.8 \\
 & SigLIP & 28.6 & 23.0 & 31.0 & 38.5 \\
 & mxbai & 71.9 & 71.0 & 71.8 & 74.4 \\
& \multicolumn{1}{l}{\textbf{\method}} \\
 & CLIP & 54.7 & 67.8 & 59.3 & 12.1 \\
 & QQMM-v2 & 72.7 & 68.0 & 77.5 & 76.3 \\
 & Seed1.6-Embed & 65.4 & 71.3 & 67.8 & 45.5 \\
 & SigLIP & 58.3 & \textbf{72.4} & 64.4 & 11.8 \\
 & \textbf{VLM Only} & \textbf{74.8} & 70.0 & \textbf{76.1} & \textbf{84.6} \\

\cmidrule(lr){1-2}
& \multicolumn{1}{l}{\textbf{\remembr}} \\
\multirow{10}{*}{\textbf{GPT-5.2}}
 & CLIP & 22.4 & 17.0 & 28.2 & 25.6 \\
 & QQMM-v2 & 35.2 & 22.0 & 46.5 & 48.7 \\
 & Seed1.6-Embed & 32.9 & 18.0 & 45.1 & 48.7 \\
 & SigLIP & 24.8 & 21.0 & 28.2 & 28.2 \\
 & mxbai & 66.7 & \textbf{70.0} & 64.8 & 61.5 \\
& \multicolumn{1}{l}{\textbf{\method}} \\
 & CLIP & 54.5 & 55.6 & 47.9 & 64.1 \\
 & QQMM-v2 & \textbf{67.9} & 63.6 & \textbf{70.4} & 74.4 \\
 & Seed1.6-Embed & 63.2 & 61.6 & 60.6 & 71.8 \\
 & SigLIP & 64.1 & 62.6 & 64.8 & 66.7 \\
 & \textbf{VLM Only} & 66.7 & 60.0 & 64.8 & \textbf{87.2} \\

\end{longtable}
}

%% file: fig_inputs/appx_figs_input/fig_simulation_example.tex
\begin{figure}[ht]
  \centering
  \includegraphics[width=\linewidth]{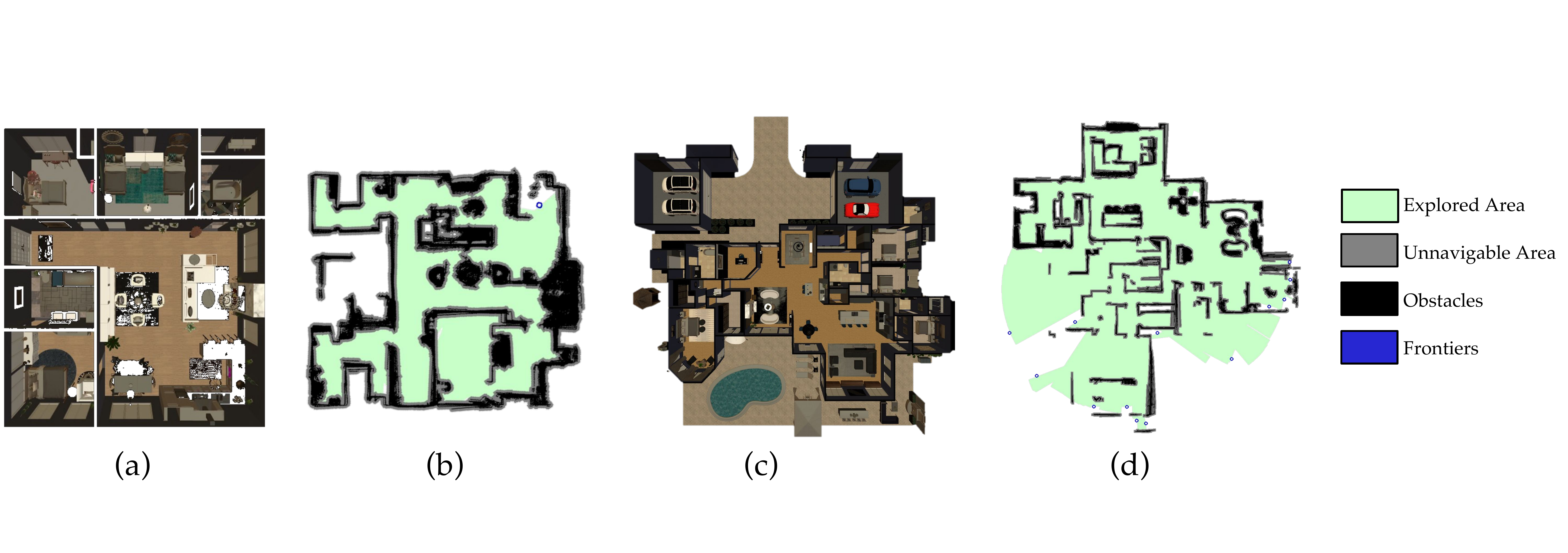}
  \caption{(a) and (c): topdown view of two example scenes from habitat simulation; (b) and (d): 2D occupancy map generated from the exploration of (a) and (c).}
  \label{fig:simulation_example}
\end{figure}

%% file: tabs/darpa_sim_table_full.tex

\begin{table*}[t]
\centering
\caption{Per-Category Retrieval Results on the Habitat Simulation Episodes of \ourdataset (3 Seeds)}
\label{tab:darpa_sim_full}
\footnotesize
\setlength{\tabcolsep}{2.2pt}
\renewcommand{\arraystretch}{1.05}
\resizebox{\textwidth}{!}{%
\begin{tabular}{lll ccc cc c c}
\toprule
& 3 seeds & Scene1--19 &
\multicolumn{3}{c}{Close VLMs (topk=5)} &
\multicolumn{2}{c}{Open VLMs (topk=1)} &
Embedder Only & Random \\
\cmidrule(lr){4-6}\cmidrule(lr){7-8}
& & mean/std &
Gemini-2.5-Flash & GPT-5.2 & Gemini-3-Pro &
Gemma3-27b & Qwen3-VL-32b &
& \\
\midrule

\multirow{12}{*}{Embedder}
& \multirow{4}{*}{QQMM-v2} & overall
& 80.47 $\pm$ 0.37 & 75.58 $\pm$ 1.95 & 86.62 $\pm$ 0.00
& 54.99 $\pm$ 3.51 & 26.11 $\pm$ 2.21
& 61.78
& \multirow{20}{*}{10.92} \\
& & reasoning
& 74.22 $\pm$ 2.04 & 66.22 $\pm$ 4.07 & 84.44 $\pm$ 0.77
& 50.67 $\pm$ 8.33 & 27.11 $\pm$ 3.08
& 53.33
& \\
& & dominant
& 88.24 $\pm$ 0.00 & 89.22 $\pm$ 3.40 & 90.20 $\pm$ 1.70
& 80.39 $\pm$ 1.70 & 36.27 $\pm$ 7.40
& 67.65
& \\
& & secondary
& 83.70 $\pm$ 2.57 & 79.26 $\pm$ 2.57 & 86.67 $\pm$ 0.00
& 40.74 $\pm$ 2.57 & 17.04 $\pm$ 2.57
& 68.89
& \\
\cmidrule{2-9}

& \multirow{4}{*}{Seed} & overall
& \textbf{81.74 $\pm$ 0.37} & 76.22 $\pm$ 1.60 & 86.84 $\pm$ 0.74
& \textbf{60.30 $\pm$ 0.97} & \textbf{27.39 $\pm$ 0.64}
& \textbf{62.42}
& \\
& & reasoning
& 75.56 $\pm$ 2.04 & 67.56 $\pm$ 1.54 & 83.11 $\pm$ 0.77
& 57.78 $\pm$ 1.54 & 24.44 $\pm$ 3.36
& 50.67
& \\
& & dominant
& 92.16 $\pm$ 1.70 & 89.22 $\pm$ 4.49 & 94.12 $\pm$ 0.00
& 82.35 $\pm$ 0.00 & 42.16 $\pm$ 6.12
& 82.35
& \\
& & secondary
& 82.96 $\pm$ 3.39 & 79.26 $\pm$ 2.57 & 86.67 $\pm$ 2.22
& 47.41 $\pm$ 1.28 & 20.74 $\pm$ 2.57
& 64.44
& \\
\cmidrule{2-9}

& \multirow{4}{*}{SigLIP} & overall
& 81.10 $\pm$ 2.24 & 71.97 $\pm$ 0.64 & 86.20 $\pm$ 2.05
& 56.05 $\pm$ 2.30 & 22.51 $\pm$ 3.89
& 61.78
& \\
& & reasoning
& 72.44 $\pm$ 3.36 & 63.11 $\pm$ 1.54 & 82.22 $\pm$ 2.78
& 49.33 $\pm$ 1.33 & 18.67 $\pm$ 2.67
& 46.67
& \\
& & dominant
& 93.14 $\pm$ 1.70 & 90.19 $\pm$ 1.70 & 90.20 $\pm$ 1.70
& 76.47 $\pm$ 2.94 & 36.27 $\pm$ 7.40
& 88.24
& \\
& & secondary
& 87.41 $\pm$ 1.28 & 71.85 $\pm$ 3.39 & 89.63 $\pm$ 1.28
& 51.85 $\pm$ 3.39 & 17.78 $\pm$ 3.85
& 66.67
& \\
\cmidrule{1-9}

& \multirow{4}{*}{VLM Only} & overall
& 76.65 $\pm$ 1.33 & \best{80.47 $\pm$ 2.05} & \textbf{88.75 $\pm$ 1.33}
& 11.04 $\pm$ 0.97 & 9.77 $\pm$ 0.37
&  & \\
& & reasoning
& 70.22 $\pm$ 2.04 & 74.22 $\pm$ 1.54 & 84.44 $\pm$ 2.04
& 12.89 $\pm$ 0.77 & 12.89 $\pm$ 0.77
&  & \\
& & dominant
& 82.35 $\pm$ 2.94 & 89.22 $\pm$ 1.70 & 92.16 $\pm$ 1.70
& 7.84 $\pm$ 1.70 & 5.88 $\pm$ 2.94
&  & \\
& & secondary
& 82.96 $\pm$ 1.28 & 82.96 $\pm$ 3.39 & 92.59 $\pm$ 1.28
& 11.11 $\pm$ 2.22 & 8.15 $\pm$ 1.28
&  & \\
\cmidrule{2-8}

& \multirow{4}{*}{ReMEmbR (QQMM-v2)} & overall
& 54.99 $\pm$ 3.51 & 57.32 $\pm$ 1.27 & 61.36 $\pm$ 0.37
& 15.71 $\pm$ 4.10 & 20.38 $\pm$ 1.10
&  & \\
& & reasoning
& 52.00 $\pm$ 6.67 & 55.56 $\pm$ 2.78 & 62.67 $\pm$ 1.33
& 15.56 $\pm$ 1.54 & 20.44 $\pm$ 0.77
&  & \\
& & dominant
& 55.88 $\pm$ 0.00 & 56.86 $\pm$ 3.40 & 57.84 $\pm$ 1.70
& 15.69 $\pm$ 7.40 & 16.67 $\pm$ 4.49
&  & \\
& & secondary
& 60.00 $\pm$ 0.00 & 60.74 $\pm$ 1.28 & 62.96 $\pm$ 2.57
& 14.81 $\pm$ 7.80 & 20.74 $\pm$ 3.39
&  & \\
\bottomrule
\end{tabular}
}
\end{table*}

%% file: tabs/navqa_table.tex
\begin{table*}[t]
\centering

\caption{Performance of \method, \remembr and \vlmonly on the NaVQA benchmark~\cite{anwar2024remembrbuildingreasoninglonghorizon} using Gemini-3-Pro. \method using the QQMM-v2 significantly outperforms all baselines in terms of overall description accuracy and localization error.}
\label{tab:navqa_results}
\setlength{\tabcolsep}{3pt}
\footnotesize
\renewcommand{\arraystretch}{1.15}

\begin{tabular}{>{\raggedright\arraybackslash}p{1.5cm}
                >{\raggedright\arraybackslash}p{1.8cm}
                ccc ccc ccc}
\toprule

\multirow{2}{*}{\textbf{Method}} &
\multirow{2}{*}{\textbf{Encoder}} &
\multicolumn{3}{c}{\textbf{$\uparrow$ Descriptive Accuracy (\%)}} &
\multicolumn{3}{c}{\textbf{$\downarrow$ Positional Error (m)}} &
\multicolumn{3}{c}{\textbf{$\downarrow$ Temporal Error (min)}} \\
\cmidrule(lr){3-5}\cmidrule(lr){6-8}\cmidrule(lr){9-11}
& & \textbf{Short} & \textbf{Medium} & \textbf{Long}
  & \textbf{Short} & \textbf{Medium} & \textbf{Long}
  & \textbf{Short} & \textbf{Medium} & \textbf{Long} \\
\midrule

\multirow{4}{*}{\textbf{\method}} & \textbf{QQMM-v2}
& \textbf{76.20} & 52.60 & \textbf{80.00}
& \textbf{2.37} & \textbf{11.48} & 38.81
& 0.23 & 0.88 & \textbf{0.07} \\

& Seed1.6-Embed
& \textbf{76.20} & 52.60 & 75.00
& 4.58 & 17.74 & 42.63
& 0.20 & 0.78 & 0.43 \\

& SigLIP
& 61.90 & 57.90 & \textbf{80.00}
& 6.51 & 14.09 & 33.20
& 0.16 & 0.81 & 1.17 \\

& CLIP-B
& 66.70 & 57.90 & 70.00
& 7.15 & 16.21 & 43.86
& 0.09 & 0.79 & 1.19 \\

\midrule
\textbf{\vlmonly} & --
& 66.70 & 63.20 & \textbf{80.00}
& 4.81 & 11.87 & \textbf{27.10}
& 0.21 & 0.73 & 0.15 \\

\midrule
\multirow{2}{*}{\textbf{\remembr}} & MixedBread
& 66.70 & \textbf{73.70} & 75.00
& 2.78 & 24.52 & 43.00
& \textbf{0.04} & 1.08 & 1.22 \\

& QQMM-v2
& 57.10 & 60.50 & 65.00
& 7.64 & 22.04 & 35.41
& 0.09 & \textbf{0.67} & 1.81 \\

\bottomrule
\end{tabular}
\end{table*}

%% file: tabs/irs_table.tex
\begin{table*}[t]
\centering
\caption{Overall system accuracy on the human-ego splits of \ourdataset (internet \& self-collected).}
\label{tab:IRS_simple_hard}

\scriptsize
\setlength{\tabcolsep}{1.0pt}
\renewcommand{\arraystretch}{1.02}

\begin{adjustbox}{max width=\textwidth}
\begin{tabular}{ll*{11}{c}}
\toprule
\multicolumn{2}{c}{} &
\multicolumn{5}{c}{\textbf{Closed VLMs} (\textbf{Gp/Gf:} Gemini Pro/Flash)} &
\multicolumn{5}{c}{\textbf{Open VLMs} (\textbf{Gm:} Gemma; \textbf{Q:} Qwen-VL)} &
\multicolumn{1}{c}{Embedder} \\
\cmidrule(lr){3-7}\cmidrule(lr){8-12}
\multicolumn{2}{c}{} &
\textbf{Gp3} & \textbf{Gp2.5} & \textbf{Gf2.5} & \textbf{GPT-5} & \textbf{GPT-4o} &
\textbf{Gm3-27b} & \textbf{Gm3-12b} & \textbf{Q2.5-32b} & \textbf{Q2.5-7b} & \textbf{Q3-32b} &
Only \\
\midrule
\midrule
\multicolumn{13}{c}{\textbf{Simple Queries} (Human-ego internet videos; 54 queries)} \\
\midrule

\textbf{Closed Embedder} & Seed1.6-Embed
 & \best{100.0} & 96.3 & \best{98.1} & \best{98.1} & \best{98.1}
 & 74.1 & \best{87.0} & 83.3 & 31.5 & 83.3
 & 85.2 \\

\cmidrule{1-13}

\multirow{3}{*}{\textbf{Open Embedders}}
& QQMM-v2
 & 98.1 & \best{98.1} & 96.3 & 96.3 & 96.3
 & \best{85.2} & 81.5 & \best{92.6} & 22.2 & \best{92.6}
 & \best{90.7} \\
& SigLIP
 & 98.1 & \best{98.1} & 94.4 & 94.4 & 94.4
 & 75.9 & 75.9 & 79.6 & 31.5 & 72.2
 & 83.3 \\
& CLIP-B
 & 94.4 & 96.3 & 94.4 & \best{98.1} & 88.9
 & 77.8 & 74.1 & 59.3 & 18.5 & 72.2
 & 74.1 \\

\cmidrule{1-13}

& VLM Only
 & \best{100.0} & \best{98.1} & 96.3 & 83.3 & 94.4
 & 13.0 & 14.8 & 29.6 & 13.0 & 25.9
 & N/A \\
& Caption (4o-mini)
 & 87.0 & 96.3 & 92.6 & 90.7 & 88.9
 & 18.5 & 27.8 & 68.5 & \best{59.3} & 90.7
 & N/A \\

\midrule
\midrule
\multicolumn{13}{c}{\textbf{Hard Queries} (Self-collected human-ego videos; 41 queries)} \\
\midrule

\textbf{Closed Embedder} & Seed1.6-Embed
 & 95.1 & 78.0 & 80.5 & \best{82.9} & 73.2
 & 53.7 & 34.1 & \best{51.2} & 4.9 & 58.5
 & \best{63.4} \\

\cmidrule{1-13}

\multirow{3}{*}{\textbf{Open Embedders}}
& QQMM-v2
 & 92.7 & \best{82.9} & \best{82.9} & \best{82.9} & 73.2
 & 51.2 & \best{43.9} & 46.3 & 2.4 & 48.8
 & \best{63.4} \\
& SigLIP
 & 85.4 & 80.5 & 68.3 & 80.5 & 73.2
 & \best{56.1} & \best{43.9} & 48.8 & 2.4 & 41.5
 & 61.0 \\
& CLIP-B
 & 82.9 & 63.4 & 56.1 & 73.2 & 56.1
 & 46.3 & 24.4 & 26.8 & 2.4 & 34.1
 & 39.0 \\

\cmidrule{1-13}

& VLM Only
 & \best{96.8} & \best{82.9} & 78.0 & 80.5 & \best{85.4}
 & 24.4 & 19.5 & 36.6 & 7.3 & 14.6
 & N/A \\
& Caption (4o-mini)
 & 68.3 & 58.5 & 58.5 & 61.0 & 56.1
 & 22.0 & 17.1 & 41.5 & \best{17.1} & \best{63.4}
 & N/A \\

\bottomrule
\end{tabular}
\end{adjustbox}
\end{table*}

%% file: fig_inputs/fig_robot.tex
\begin{figure}[t]
  \centering
  \includegraphics[width=0.3\linewidth]{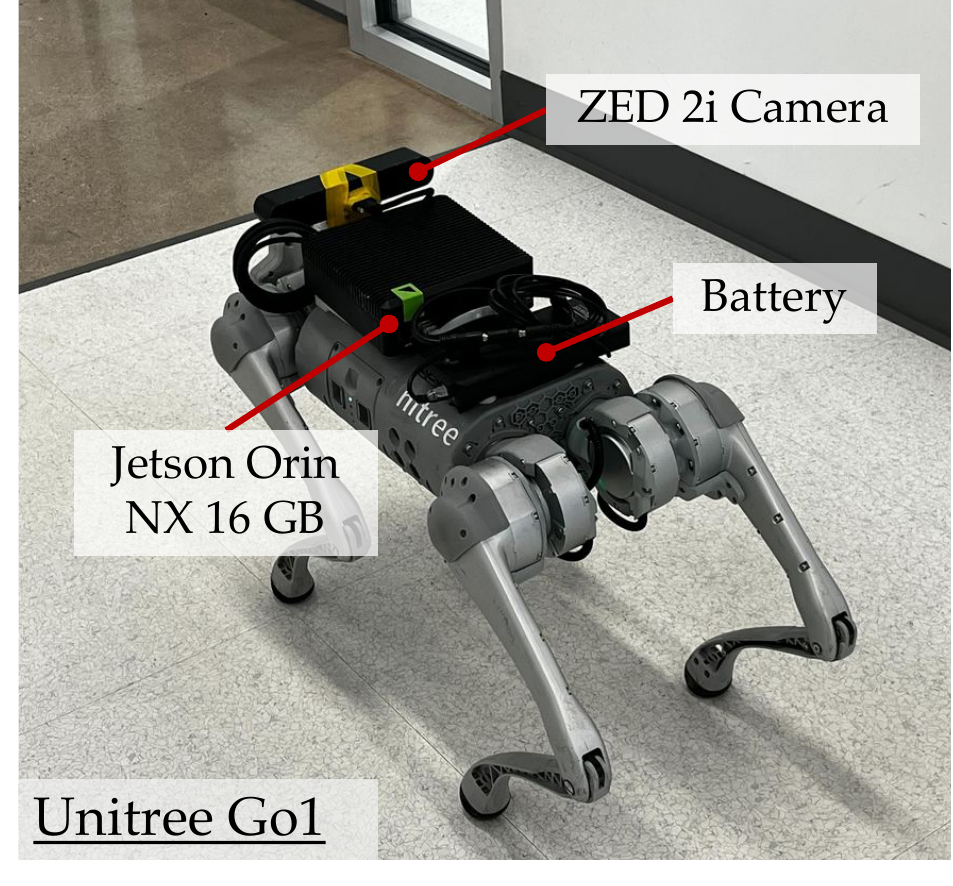}
  \caption{The real-world robot configuration. Unitree Go1 has a ZED 2i camera to provide RGB-D and point cloud information via SLAM. We use Jetson Orin as the edge computer for this robot. (Appendix~\ref{appx_short:results_and_implement_details})}
  \label{fig:robot}
\end{figure}

%% file: fig_inputs/appx_figs_input/fig_frontier_exploration.tex
\begin{figure}[ht]
  \centering
  \includegraphics[width=0.55\linewidth]{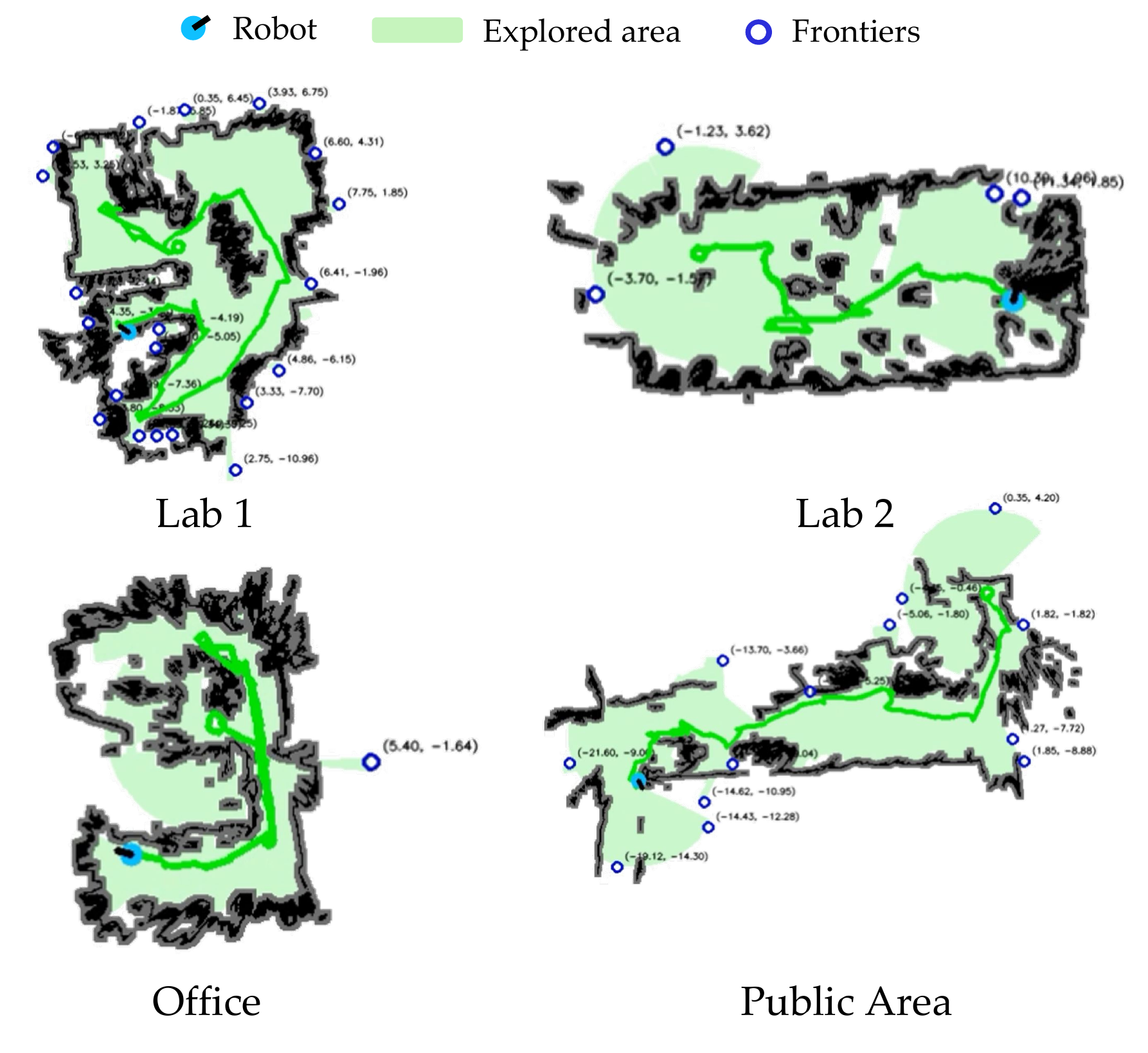}
  \caption{Four 2D occupancy map generated from the exploration of Lab 1, Lab 2, Office and Public Area. Dark green line means the trajectory of the robot when it navigation these indoor scenes during exploration.}
  \label{fig:frontier_exploration}
\end{figure}

%% file: fig_inputs/appx_figs_input/fig_waypoints.tex
\begin{figure}[ht]
  \centering
  \includegraphics[width=\linewidth]{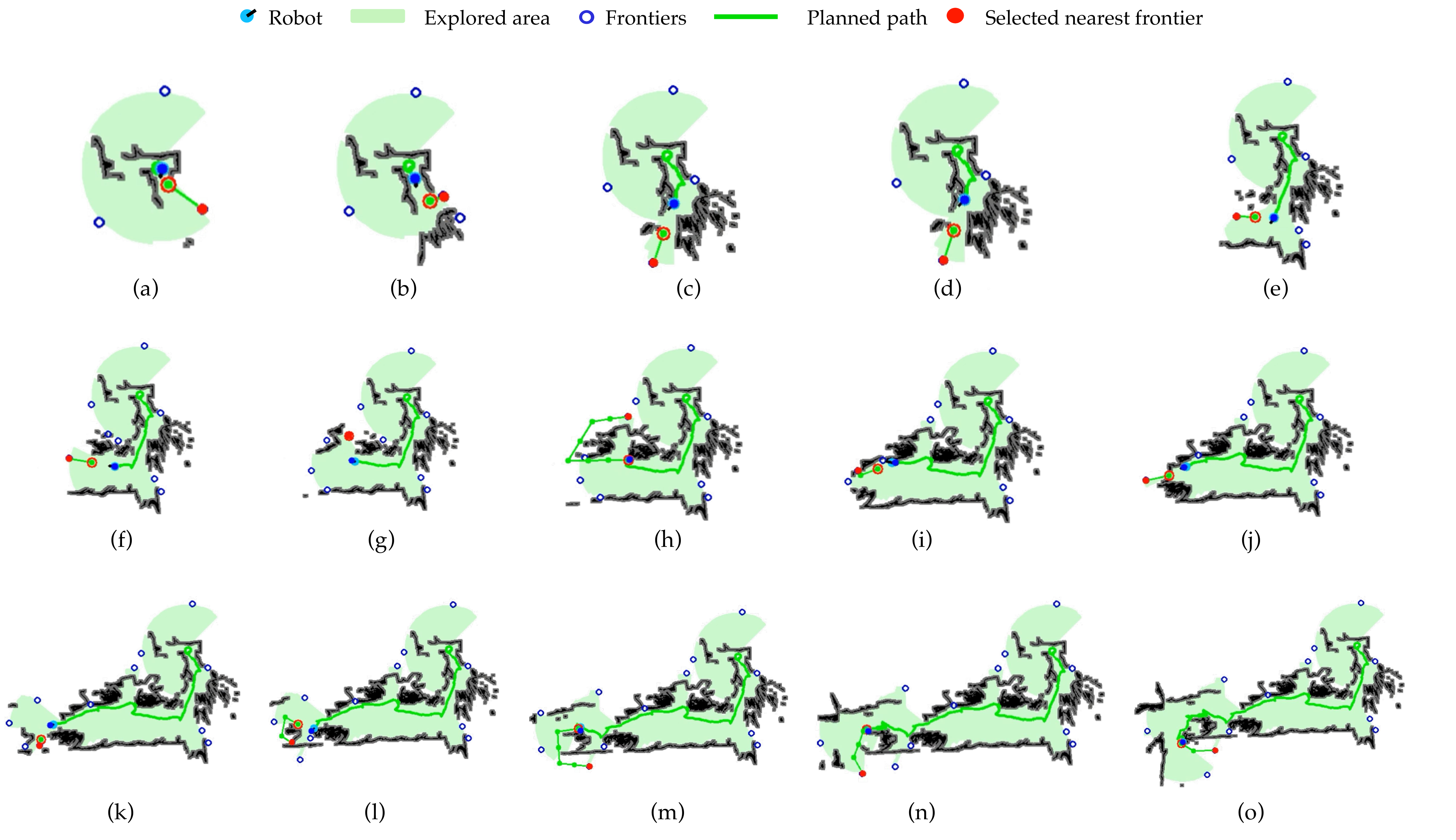}
  \caption{Dynamic visualization of frontier-based exploration in an unseen environment.
Panels (a)–(o) show successive timesteps during exploration.
At each iteration, the robot plans a path (dark green) toward the nearest selected frontier (blue) on the boundary between explored and unexplored space, expands the explored region (light green), and updates the occupancy map accordingly.
Red dots denote the robot’s pose over time.
The sequence demonstrates the iterative frontier selection and execution process used in our exploration pipeline.}
  \label{fig:waypoints}
\end{figure}

%% file: fig_inputs/appx_figs_input/fig_retireval_bowen_vlm_only.tex
\begin{figure}[ht]
  \centering
  \includegraphics[width=0.8\linewidth]{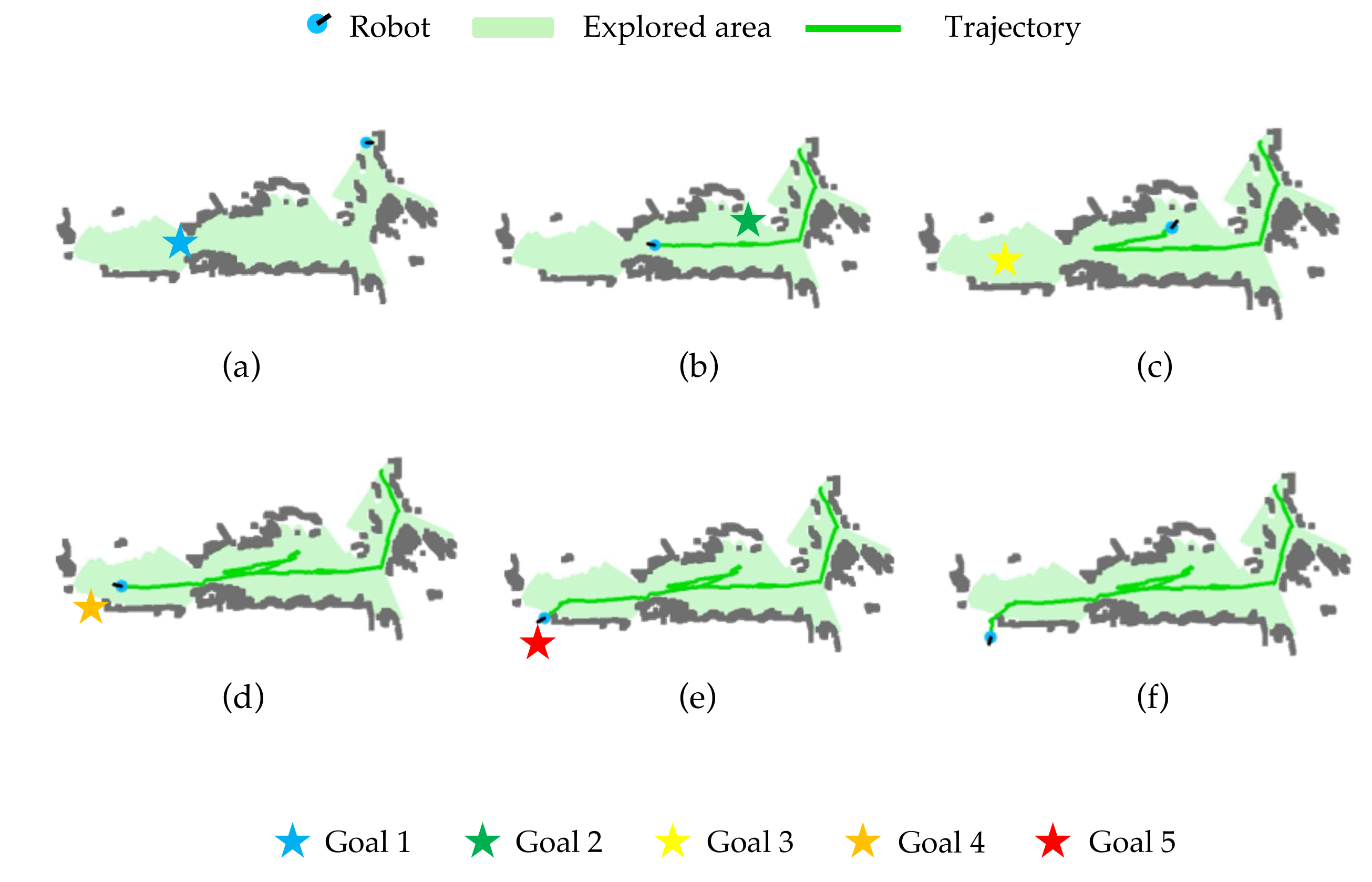}
  \caption{Visualization of sequential goal navigation on a built map during the execution phase.
Panels (a)--(f) show an example episode in which the robot navigates to five goal locations in order (colored stars).
At each step, the robot’s current pose is shown in blue, the explored area is shaded in light green, and the executed trajectory is overlaid in green.
The sequence illustrates how the robot reuses the accumulated map to plan and execute goal-directed navigation across multiple targets.}

  \label{fig:retireval_bowen_vlm_only}
\end{figure}

%% file: tabs/appx_tabs/params.tex
\begin{table*}[ht]
\centering
\caption{Hyper-Parameter Configurations of Our Benchmark Evaluation}
\label{tab:appx_hyperparams_list}
\scriptsize
\setlength{\tabcolsep}{4.0pt}
\renewcommand{\arraystretch}{1.15}
\begin{tabular}{lccccccc}
\toprule
& \multicolumn{3}{c}{\textbf{closed models}} & \multicolumn{3}{c}{\textbf{open models}} \\
\cmidrule(lr){2-4} \cmidrule(lr){5-7}
\textbf{Benchmark} & \textbf{Top-K} & \textbf{\#Max Tool Call} & \textbf{Temperature} & \textbf{Top-K} & \textbf{\#Max Tool Call} & \textbf{Temperature} \\
\midrule
RAVEN-QA Human Ego (Web+Ours) & 5 & 3 & 0.0001 & 5 & 3 & 0.0001 \\
RAVEN-QA Robot Ego (Sim+Real) & 5 & 3 & 0.6 & 1 & 3 & 0.5 \\
FindingDory & Time 200, Text 30, Position 1 & 15 & 0.7 & - & - & - \\
NaVQA & 5 & 3 & 0.7 & - & - & - \\
\bottomrule
\end{tabular}
\end{table*}

%% file: fig_inputs/appx_figs_input/fig_pilot_study.tex
\begin{figure}[ht]
  \centering
  \includegraphics[width=.8\linewidth]{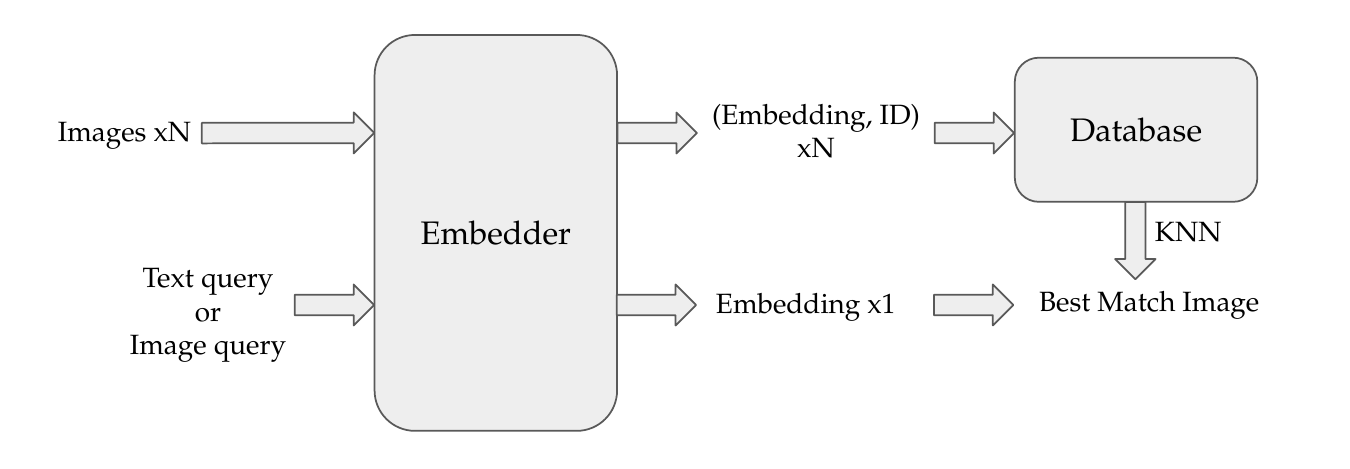}
  \caption{The pipeline of the problem setup in our Image-query and Text-query Image Reverse Search study (I-IRS and T-IRS).}
  \label{fig:appx_pilot_study}
\end{figure}

%% file: fig_inputs/appx_figs_input/fig_irs_query_examples.tex

%% file: tabs/appx_tabs/pilot_study_table_text_q.tex


\begin{table*}[ht]
\centering
\caption{Leaderboard: Model Specifications and Accuracy across Text and Image Queries. \emph{Dom} is the accuracy on the dominant object retrieval, and \emph{Sec} is the accuracy on the secondary object retrieval.}
\label{tab:merged_pilot_accuracy_results}
\footnotesize
\begin{tabular}{llc|ccc|c}
\toprule
\multirow{2}{*}{\textbf{Model}} & \multirow{2}{*}{\textbf{Group}} & \multirow{2}{*}{\textbf{Size}} & \multicolumn{3}{c|}{\textbf{Text-Query}} & \textbf{Image-Query} \\
\cmidrule(lr){4-6}
& & & \textbf{Overall(\%)} & \textbf{Dom.} & \textbf{Sec.} & \textbf{Overall (\%)} \\
\midrule
QQMM-Embed-v2 & A: Contrastive loss & 8.3B & 91.8 & 100 & 82.6 & 89.8 \\
Seed-1.6 & D: Closed-source & Unknown & 85.7 & 100 & 69.6 & 87.8 \\
SigLIP-so400M-384 & A: Contrastive loss & 880M & 81.6 & 95.2 & 65.2 & 55.1 \\
Google MM Embedding & D: Closed-source & Unknown & 79.6 & 100 & 60.9 & 46.9 \\
CLIP-ViT-H-14 (1B) & A: Contrastive loss & 1B & 79.6 & 95.2 & 60.9 & 69.4 \\
CLIP-ViT-B-32 & A: Contrastive loss & 150M & 73.5 & 95.2 & 56.5 & 57.1 \\
DinoV3 (ViT-7B-16) & C: Self-distillation & 7B & --- & --- & --- & 77.6 \\
DinoV3 (ViT-S-16) & C: Self-distillation & 22M & --- & --- & --- & 63.3 \\
PaliGemma2-3B & B: Autoregressive & 3B & 51.0 & 66.7 & 43.5 & 67.4 \\
Qwen-2.5VL-32B & B: Autoregressive & 32B & --- & --- & --- & 67.4 \\
Qwen-2.5VL-3B & B: Autoregressive & 3B & 14.3 & 9.5 & 17.4 & 55.1 \\
Random Select & E: Random Baseline & --- & 5.5 & 5.5 & 5.5 & 5.5 \\
\bottomrule
\end{tabular}
\end{table*}

%% file: tabs/appx_tabs/pilot_study_table_image_q.tex


%% file: tabs/appx_tabs/pilot_study_top_k_closer_look.tex
\begin{table*}[ht]
\centering
\caption{An overview of top-K retrieval performance under text and image queries. We report the average Top-2 similarity ratio for correct cases (A), the average index of ground-truth in Top-$K$ for incorrect cases (B), and overall accuracy (C). List them in A/B/C format.}
\label{tab:appx_embedding_closer_look}
\footnotesize
\setlength{\tabcolsep}{6pt}
\renewcommand{\arraystretch}{1.15}
\begin{tabular}{lcc}
\toprule
\textbf{Model} &
\textbf{Text Query} &
\textbf{Image Query} \\
\midrule
QQMM-Embed-v2 (8B, Contrastive Loss) &
\textbf{1.98 / 3.5 / 92} &
\textbf{1.65 / 3.4 / 90} \\
CLIP (ViT-H-14) (1B, Contrastive Loss) &
1.38 / 4.0 / 80 &
1.20 / 4.4 / 69 \\
Qwen2.5-VL (3B, Autoregressive Loss) &
1.02 / 8.1 / 14 &
1.04 / 4.3 / 55 \\
\bottomrule
\end{tabular}
\end{table*}

%% file: fig_inputs/appx_figs_input/framework_study_topk_sim_scores.tex
\begin{figure}[ht]
  \centering
  \includegraphics[width=.65\linewidth]{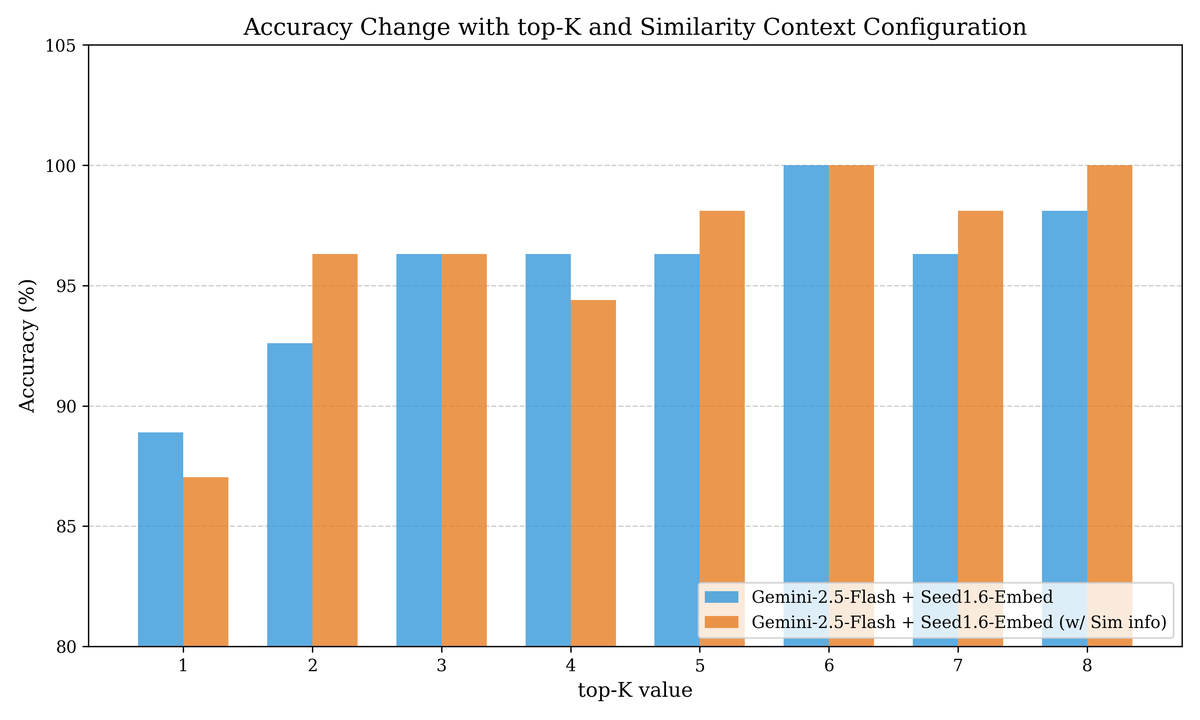}
  \caption{We observed that for the Simple split, retrieving a reasonably larger top-K memory set and providing the VLM agent with similarity scores yield performance gains.}
  \label{fig:appx_topk_simi_score}
\end{figure}

%% file: fig_inputs/appx_figs_input/fig_ordering_intuition.tex
\begin{figure}[ht]
  \centering
  \includegraphics[width=\linewidth]{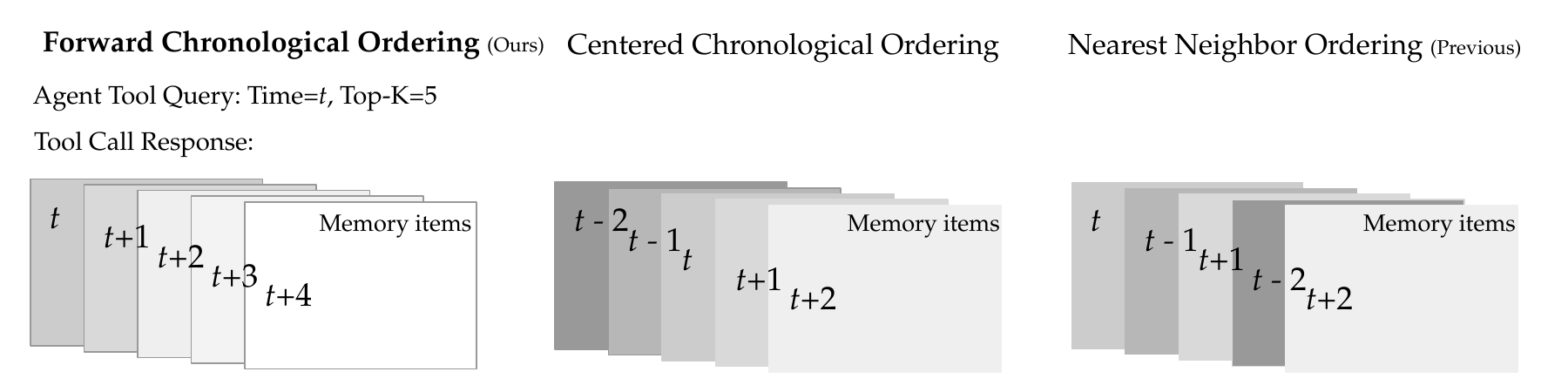}
  \caption{The intuition and nuances of the ordering methods in time-based retrieval call response.}
  \label{fig:appx_retrieval_ordering_intuition}
\end{figure}

%% file: tabs/appx_tabs/ordering_exp_results.tex
\begin{table}[ht]
\centering
\caption{Comparison of FindingDory (ep\_1) Accuracy Across Different Ordering Methods}
Base Agent=Gemini-2.5-Flash, Embedder=QQMM, Top-K=100 \\
\label{tab:appx_ordering_exp_results}
\begin{tabular}{lccc}
\toprule
\textbf{Ordering Method} & \textbf{Forward Chronological} & \textbf{Centered Chronological} & \textbf{Nearest Neighbor} \\
\midrule
Accuracy (\%) & \textbf{49.00\%} & 32.65\% & 42.86\% \\
\bottomrule
\end{tabular}
\end{table}

%% file: fig_inputs/rebuttal_fig_inputs/judgement_accuracy_vs_stride.tex
\begin{figure}
    \centering
    \includegraphics[width=.9\linewidth]{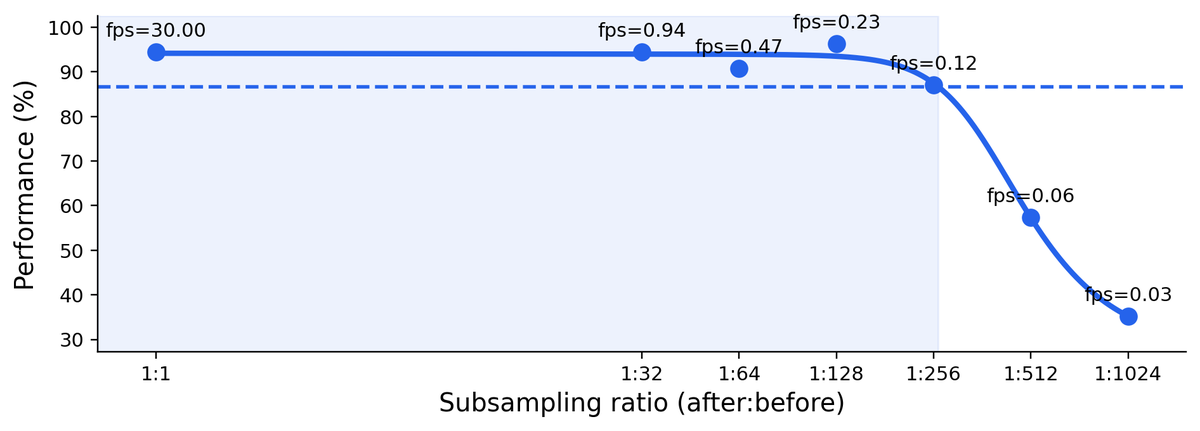}
    \caption{RAVEN memory is highly compact and scalable. Dashed line shows $10\%$ performance drop at $264.5\times$ compression.}
    \label{fig:compression}
    \vspace{-10pt}
\end{figure}

%% file: fig_inputs/rebuttal_fig_inputs/rebuttal_coverage_experiment.tex
\begin{figure}[ht]
  \centering
\includegraphics[width=\linewidth]{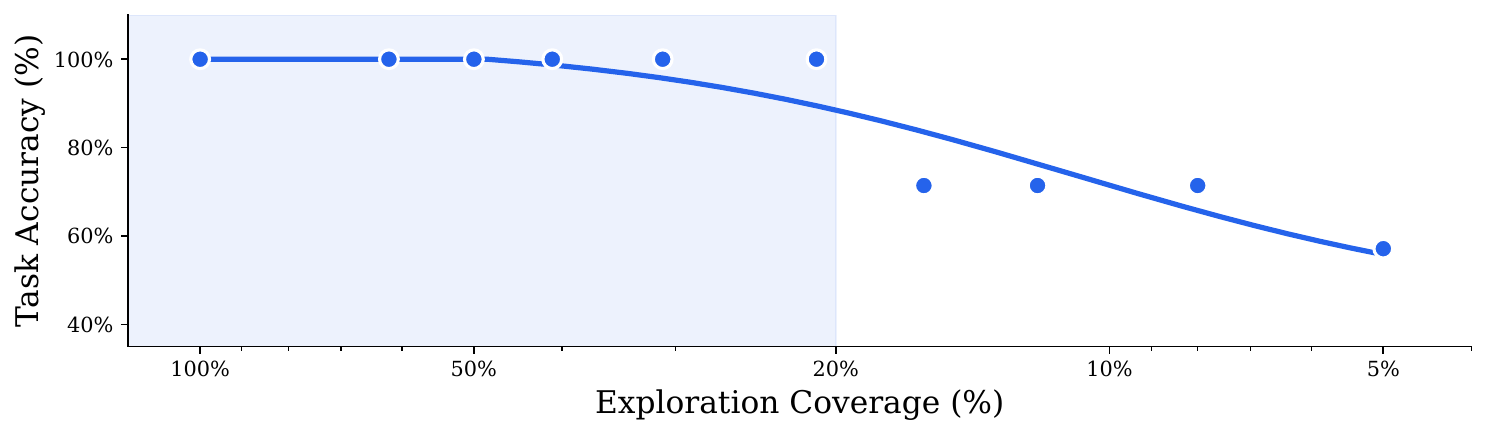}
  \caption{RAVEN performs strongly even at low exploration coverage, retrieving with $100\%$ accuracy at $21\%$ coverage.}
\vspace{-20pt}
  \label{fig:rebuttal_coverage_experiment}
\end{figure}

%% file: secs/8.C-appendix.tex
\section{Visualizations and Case Examples}
\label{appx_short:visual_and_examples}
\vspace{10pt}

\subsection{Justifications on Caption Bottleneck}
\label{appendix:caption_bottleneck}
Fig.~\ref{fig:caption_bottleneck} shows the image-to-caption loss during the caption process.
In text-based pipelines, raw visual observations are first compressed into natural language captions and then embedded into a vector database.
This process introduces a semantic bottleneck: only concepts explicitly verbalized by the captioner are retained, while fine-grained visual information such as shape, spatial layout, and implicit semantic relations is often lost.
In this chicken-thigh example, the captioner may focus only on salient attributes such as texture, pink color, and bone shape, and thus produce an underspecified description such as \emph{meat in a refrigerator}. As a result, a text-based memory pipeline can fail to retrieve the correct evidence for the query \emph{chicken thighs}, since the relevant distinction is not preserved in the caption.
In contrast, \method{} stores visual experiences directly as multimodal embeddings augmented with spatial and temporal metadata.
By avoiding intermediate captioning, the memory preserves richer visual semantics and spatial structure, enabling more reliable retrieval for queries that require fine-grained discrimination or implicit visual reasoning.

\input{fig_inputs/appx_figs_input/fig_caption_bottleneck}
Considering efficiency and cost due to the number of frames to caption, we chose GPT-4o-mini as the captioner.

\vspace{20pt}

\subsection{Failure Modes -- Under-Captioning}

Under-captioning comes into play as the main reason why \remembr suffers from \textit{information bottleneck}. We show some failure mode examples here.
Most failures are from \textbf{under-captioning}, especially for the harder queries. If the VLM does not explicitly verbalize a specific detail in the view, the agent will be highly likely to fail on retrieving that, like those easily neglectable, small items, or more fine-grained labels, e.g., "chicken thighs" versus "food" (see below).

\input{fig_inputs/appx_figs_input/fig_under_captioning}

\begin{lstlisting}[language=python,
  literate={chicken\ thigh}{{\textbf{chicken thigh}}}{13}
]
# (no chicken thighs mentioned)
caption = "You are in a grocery store, specifically in the refrigerated section. Here's a detailed description of what you see:1. **Shelving and Products**:    - There are multiple shelves filled with packaged food items. The shelves are organized with clear pricing labels.   - The top shelf has items priced between $4.26 and $4.52.\n   - The middle shelf prominently displays items priced..." 

query = "chicken thighs"
\end{lstlisting}

\vspace{20pt}

\subsection{Failure Modes -- Over-Captioning}

Some failure modes, on the other hand, comes from \textbf{over-captioning}, such as “too” detailed verbalization. For instance, for Fig.~\ref{fig:appx_over_cap}, the model captions as follows.

\input{fig_inputs/appx_figs_input/fig_over_captioning}

\begin{lstlisting}[language=python,
  literate={fireplace}{{\textbf{fireplace}}}{9}
]
caption = "The image shows a large, well-designed house with a complex roof structure. The roof is covered with dark shingles, and there are multiple gables and dormer windows, adding architectural interest. A prominent feature is a stone chimney, suggesting the presence of a fireplace inside. The house has a mix of materials, including wood and stone, giving it a rustic yet modern appearance. The windows are large, allowing for plenty of natural light, and are framed with light-colored trim that contrasts with the darker roof. The surrounding environment includes some trees, indicating a natural, possibly suburban or rural setting. The sky is clear with a few clouds, suggesting a pleasant day.\n\nOverall, the house appears spacious and well-maintained, with an emphasis on blending with the natural surroundings."
\end{lstlisting}

When the agent is given this query: "fireplace", it finally chose this image probably because the caption explicitly mentions the word "fireplace", even though there is no fireplace in this frame. In \remembr, the model cannot actually “see” the image; this nature gives rise to this seemingly unrelated, "absurd" answer.

As a reference, the groundtruth image and its caption are in Fig.~\ref{fig:appx_over_cap_gt}.

\input{fig_inputs/appx_figs_input/fig_over_captioning_gt}

\begin{lstlisting}[language=python,
  literate={fireplace}{{\textbf{fireplace}}}{9}
]
query = "fireplace"

caption = "1. **Furniture and Layout**:\n   
    - There is a floral-patterned sofa with yellow cushions, ...
    
    2. **Decor and Art**:
    - Two framed artworks hang on the walls. One is above the fireplace, and the other is above the sofa.   
    - The fireplace is white, with a mantel that holds decorative items, possibly candles or small sculptures.\n\n3. **Flooring and Rugs**:\n   
    - ...
    
    5. **Additional Features**:\n   
    - A side table next to the sofa holds ...
    ...
    
    Overall, the room exudes a comfortable and welcoming ambiance, with a mix of traditional and cozy elements."
\end{lstlisting}

We are uncertain why the model Gemini-2.5 chose the former one as its answer, but at least this failure mode shows it could be hard to control the granularity of captioning through prompts, leading to not only under-captioning, but also over-captioning, just as the former caption shows. We use the originial \remembr's prompt for captioning.

\vspace{20pt}

\subsection{\method Reasoning Example 1}
Here, we demonstrate two fully elaborated examples step-by-step showing how \method uses tools, reasons about the context, and act in the real environments, to navigate to a requested point of interest with great performance and efficiency.

We take a query of the real-world dataset in our research labs as an example here. The settings are VLM=GPT-5.2 (2025-12-11 snapshot) and Embedder=SigLIP (ViT-SO400M-14-SigLIP-384+webli+1152). Below are the in-detail reasoning steps.
The query is:
\begin{lstlisting}[language=python]
Query = "Adam is just heading to his daily sports training. He went out with only his backpack. Now he comes back and looks for his clothes. Question: Where are they?"
\end{lstlisting}

The agent starts with reasoning about using which tool:

\begin{lstlisting}[language=python]
context_reasoning = "No memories (images/time/position) have been provided yet, so I cannot know where Adam's clothes are located. I need to retrieve relevant moments showing Adam leaving/returning and where clothes were placed (e.g., bedroom, closet, chair, laundry basket). Text-based retrieval is appropriate because I don't have a specific time or coordinates." 

tool_reasoning = "Use retrieve_from_text to search the video memory for scenes mentioning/depicting Adam's clothes, changing, closet, laundry basket, or where he put them before leaving for training."

tool = "retrieve_from_text"

tool_input = "Adam clothes location before leaving for sports training closet chair laundry basket bedroom"
\end{lstlisting}

The retrieval tool responds with: Fig.~\ref{fig:appx_reasoning_examples_1}.

\input{fig_inputs/appx_figs_input/fig_reasoning_examples_1}

Given the context, the agent continues to reason the next steps:

\begin{lstlisting}[language=python]
context_reasoning = "We need locate Adam's clothes. Current retrieved frames (19:01:30, 19:05:03, 19:02:00, 19:04:02, 19:04:30) show a lab/office: chairs, desks, cabinets, fridge, tripod, coat stand with a dark jacket, boxes. No clear pile of clothes or backpack. One frame shows a jacket hanging, could be the clothes but question asks 'his clothes' likely training clothes left somewhere. Evidence insufficient; need search specifically for clothes like shirt, pants, shoes, jacket, gym clothes, laundry, changing area, bag."

tool_reasoning = "Use retrieve_from_text with a new, more targeted query for 'clothes' (shirt/pants) and potential locations like chair, floor, hanger, near door, bench. Avoid same query as before by using different wording."

tool = "retrieve_from_text"

tool_input = "clothes jacket hoodie shirt pants on coat rack coat stand hanger chair bench locker gym towel"
\end{lstlisting}

The second retrieval call returns: Fig.~\ref{fig:appx_reasoning_examples_2}.

\input{fig_inputs/appx_figs_input/fig_reasoning_examples_2}

Conditioned on the new context, the agent decided to generate a final response for robot execution.

\begin{lstlisting}[language=python]
context_reasoning = "We need location of Adam's clothes. In retrieved frames, the only clear clothing item is a black coat/garment hanging on a coat stand next to a desk/workbench. Frame at 19:04:02 and 19:04:00 shows a black garment draped over a standing coat rack beside a gray desk with computers underneath, near a hallway/open lab area. No other clothes visible. This is sufficient to answer: clothes are on the coat rack by the desk."

tool = "generate_response"    

response = "From the robot's memories, Adam's clothes are hanging on the coat rack/stand next to the desk/workbench area (seen around 19:04:00 - 19:04:02)."
\end{lstlisting}

The final response decides that a position answer should be finally delivered. Below shows the target location to navigate to.

\begin{lstlisting}[language=python]
question_type_reasoning = "The question asks 'Where are they?', which is requesting a location, so a position answer is expected."
question_type = "position"

final_response_reasoning = "From the retrieved memory images, the most relevant clue of 'clothes' is a black garment hanging on a coat stand next to a table/cabinet. That scene is recorded at position [1.71, -4.13, 0.47], so I will report that as the clothes location.", "text": "Adam's clothes are hanging on the coat stand next to the table, around position (1.71, -4.13, 0.47)"

position = "[1.71,-4.13,0.47]"

yaw = "3.14"
\end{lstlisting}

\vspace{20pt}

\subsection{\method Reasoning Example 2} Below are reasoning steps from an another example, in the same setting but at a different research place.

The query is,
\begin{lstlisting}[language=python]
Query = "My friend Alex prepared a surprise for me and said he stealthily hid something behind a cabinet. That was the gift for me if I could find it. Robot, please go back to the place where you look directly into a gap between the cabinet and the wall."
\end{lstlisting}

The agent reasons about tool-using,

\begin{lstlisting}[language=python]
context_reasoning = "No prior memory frames, times, or positions are provided. To go back to where the robot looked directly into the gap between a cabinet and the wall, I must first locate that scene in memory. The most relevant retrieval is a text search for a cabinet-wall gap viewpoint."

tool_reasoning = "Use retrieve_from_text to find memories showing a gap between a cabinet and the wall / looking behind a cabinet."

tool = "retrieve_from_text"

tool_input = "look directly into a gap between the cabinet and the wall behind a cabinet"
\end{lstlisting}

The retrieval tool returns: Fig.~\ref{fig:appx_reasoning_examples_3}.

\input{fig_inputs/appx_figs_input/fig_reasoning_examples_3}

Given the context, reasoning goes on:

\begin{lstlisting}[language=python]
context_reasoning = "Goal: identify where the robot previously looked directly into the gap between a cabinet and the wall, so it can go back there. From the 5 retrieved frames: (1) at 19:10:00 shows a close-up narrow vertical gap between a tall cabinet/appliance and an adjacent surface (likely wall) \u2014 this matches 'look directly into a gap'. (2) at 19:02:00 shows the side/back of a cabinet with open space behind but not a direct tight gap view. (3) and (5) are hallway/corner views; (4) is a room-wide shot with chairs/door; none show a direct cabinet-wall gap. Therefore we have enough: use the position from (1) to navigate back."

tool_reasoning = "We have the best matching memory frame and its (x,y,z) position, so no further retrieval is needed. Provide the target coordinates to go back to that spot."

tool = "__conversational_response"

tool_input = "The spot where I'm looking directly into the narrow gap between the cabinet and the wall is from memory at 19:10:00. Go to approximately position (-18.03, -9.85, -0.19)."
\end{lstlisting}

The final generated answer delivered to the user is

\begin{lstlisting}[language=python]
question_type_reasoning = "The user is asking me to go back to a specific place, which requires a location coordinate, so the expected answer type is position."
question_type = "position"

final_response_reasoning = "From the retrieved memories, the clearest view looking into a narrow gap between a cabinet and the wall is in the frame at 19:10:00, where the camera is aimed directly into a tight vertical gap. The robot's recorded position then was [-18.03, -9.85, -0.19], so I will return that as the target place."

text_answer = "Go to the spot at approximately (-18.03, -9.85, -0.19) where I was looking directly into the gap between the cabinet and the wall."

position = "[-18.03, -9.85, -0.19]"

yaw = "-0.92"
\end{lstlisting}

\vspace{20pt}
\subsection{Case Study between \method, \remembr, and \vlmonly in Real World}

Failure Case Study of \remembr and \vlmonly on Three Query Examples shown in Fig.~\ref{fig:real_world}.
\label{appendix:real_fail_case_study_experiment}

The experiment configuration for three methods are:
\begin{itemize}
    \item \method: Gemini-2.5-Flash as the VLM agent and QQMM-v2 as the multimodal embedder.
    \item \remembr: Gemini-2.5-Flash as the VLM agent, GPT-4o-mini as the captioner and QQMM-v2 as the text embedder.
    \item \vlmonly: Gemini-2.5-Flash only.
\end{itemize}

\paragraph{Question about safety helmet} \vlmonly fails at this, while \method addresses this correctly.

\par \textbf{\vlmonly.}
Gemini-2.5-Flash answers:
\begin{lstlisting}[]
"Your safety helmet is on the blue chair, which was seen at approximately [7.82, -0.83, 0.5]."
\end{lstlisting}
However, the safety helmet is actually located on top of the gray seating rather than on a blue chair. Moreover, the predicted coordinate $[7.82, -0.83]$ corresponds to a sports area (e.g., with basketballs and footballs) and is unrelated to the helmet. This indicates a failure in visual grounding and spatial localization. Without structured external memory,  Gemini-2.5-Flash directly search over a limited set of retrieved images (15 frames in our setup), which can lead to confusion between visually similar objects and spurious associations (e.g., misidentifying a helmet-like object as the target).
\input{fig_inputs/fig_real_world}
\paragraph{Question about home automation} \remembr returns a wrong answer. In contrast, \method still captures the key information from visual cues and correctly solves the retrieval problem.

\textbf{Question:} ``Where is the object in the scene that is a smart-home controller box and most likely related to home automation rather than laboratory training equipment?''

\textbf{Ground Truth:} Position [6.032, -1.638] (Artech Power Controller)

\textbf{Model Answer:} Position [2.45, 0.08] (black electronic device on table)

\textbf{1. Caption Description Gap.} The ground truth object (Artech Power Controller) was never explicitly described in any caption. The captions for the frame containing the correct answer described generic objects without mentioning ``smart-home,'' ``controller,'' or ``home automation.'' This creates an insurmountable retrieval barrier for text-based methods.

\textbf{2. Retrieval Failure.} The model performed three retrieval attempts with different queries:
\begin{itemize}
    \item \texttt{smart-home controller box} - No relevant frames retrieved
    \item \texttt{home automation equipment} - No relevant frames retrieved  
    \item \texttt{control unit} - No relevant frames retrieved
\end{itemize}

None of the 15 frames' captions contained sufficient semantic similarity to these queries to surface the correct frame in the top-k results.

\textbf{3. Best-Effort Reasoning on Wrong Candidates.} Given the failed retrieval, the model attempted to reason over the available (incorrect) candidate frames. The model's reasoning process is shown below:

\begin{lstlisting}
answer_reasoning = "I have not explicitly identified an object labeled as a 'smart-home controller box' or 'home automation equipment' in my observations. However, at time 1969-12-31 19:02:30, I observed a 'black device that could be a computer or another electronic gadget' on a table. This is the most plausible candidate for a generic electronic control unit that could be related to home automation, given the lack of more specific information. I will use the robot's average position during that observation as the location for this device and provide a default orientation."
\end{lstlisting}

The model correctly acknowledged uncertainty (``I haven't seen an object explicitly described as a smart-home controller box'') but was forced to provide a best-guess answer from the wrong candidate pool.

\textbf{4. Semantic Confusion.} The retrieved captions contained descriptions of:
\begin{itemize}
    \item ``white box with a dog-like figure'' (prototype/toy)
    \item ``robotic device with wheels'' (educational tool)
    \item ``printer or scanner'' (office equipment)
    \item ``black device that could be a computer'' (selected answer)
\end{itemize}

The model reasoned that ``black electronic device'' was the closest match to ``home automation controller'' among these candidates, demonstrating reasonable inference but fundamentally limited by the retrieval stage failure.


%% file: fig_inputs/appx_figs_input/fig_caption_bottleneck.tex
\begin{figure}[htbp!]
  \centering
  \includegraphics[width=0.6\linewidth]{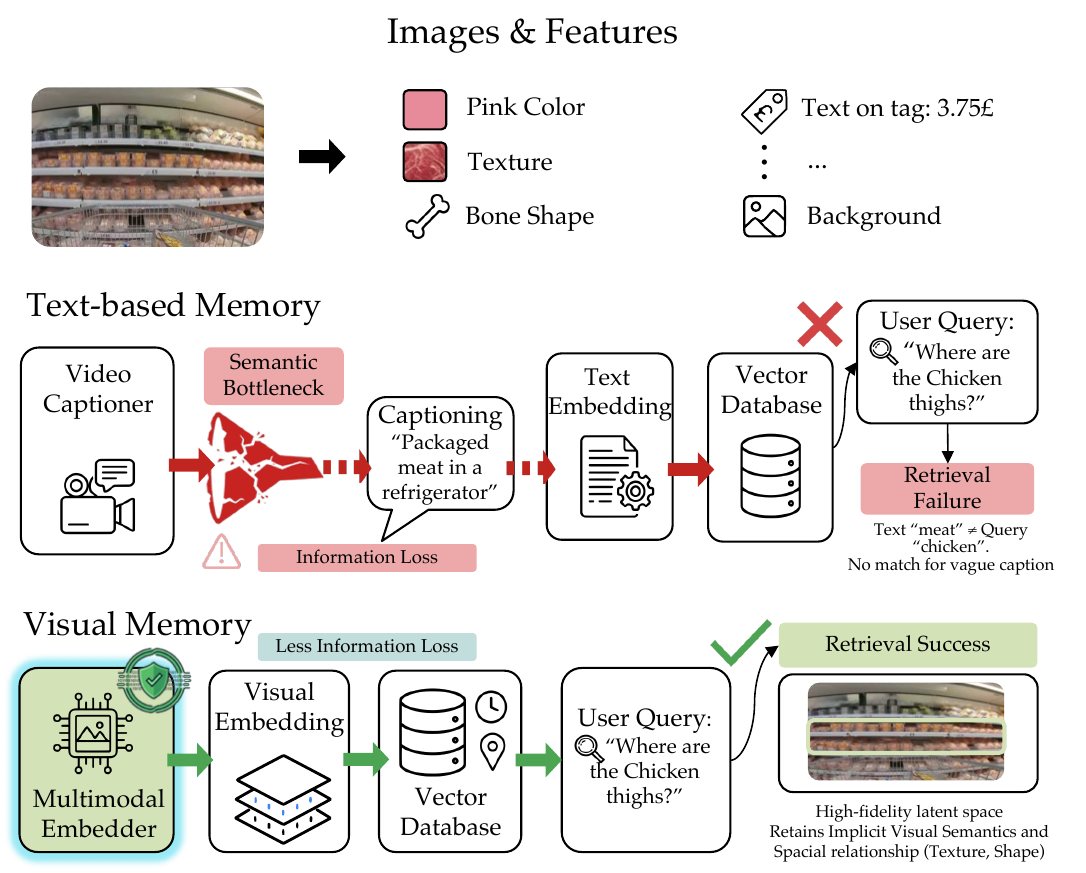}
  \caption{Comparison between text-based memory and visual embedding memory.
This figure contrasts conventional text-based memory pipelines with the proposed visual embedding memory used in \method{}. 
Top: chicken thigh figure and features captioner may extract.
Middle: text-based memory relies on video captioning followed by text embeddings, which introduces a semantic bottleneck and can discard fine-grained visual cues such as shape, and spatial relations, leading to retrieval failures.
Bottom: visual embedding memory stores multimodal embeddings directly, preserving high-fidelity visual semantics and spatial structure, enabling robust retrieval for fine-grained queries.}
  \label{fig:caption_bottleneck}
\end{figure}

%% file: fig_inputs/appx_figs_input/fig_under_captioning.tex
\begin{figure}[ht]
  \centering
  \includegraphics[width=.5\linewidth]{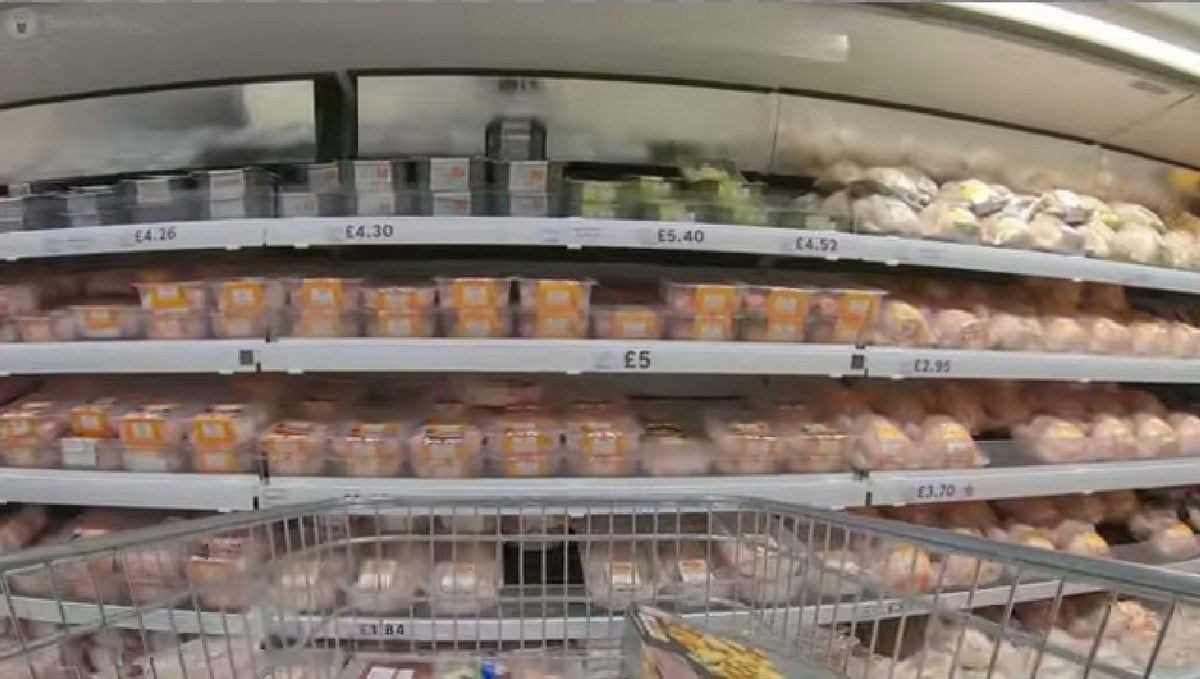}
  \caption{The groundtruth frame of the query chicken thighs, which is missed by \remembr.}
  \label{fig:appx_under_cap}
\end{figure}

%% file: fig_inputs/appx_figs_input/fig_over_captioning.tex
\begin{figure}[ht]
  \centering
  \includegraphics[width=.5\linewidth]{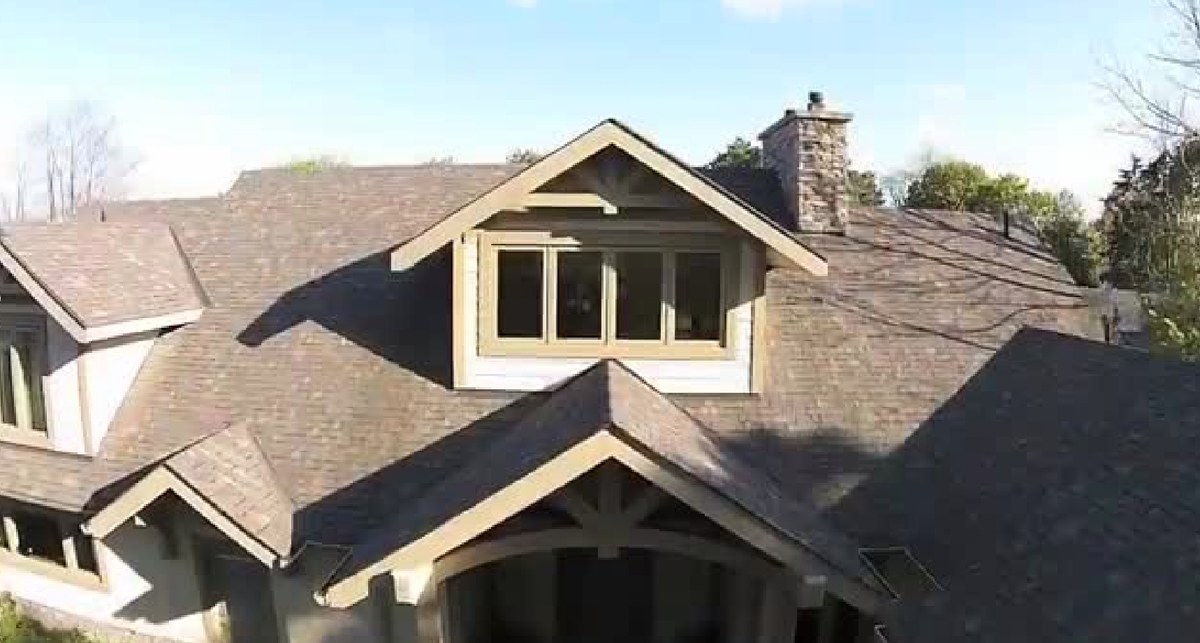}
  \caption{A wrong retrieved frame from \remembr. Query: fireplace.}
  \label{fig:appx_over_cap}
\end{figure}

%% file: fig_inputs/appx_figs_input/fig_over_captioning_gt.tex
\begin{figure}[ht]
  \centering
  \includegraphics[width=.5\linewidth]{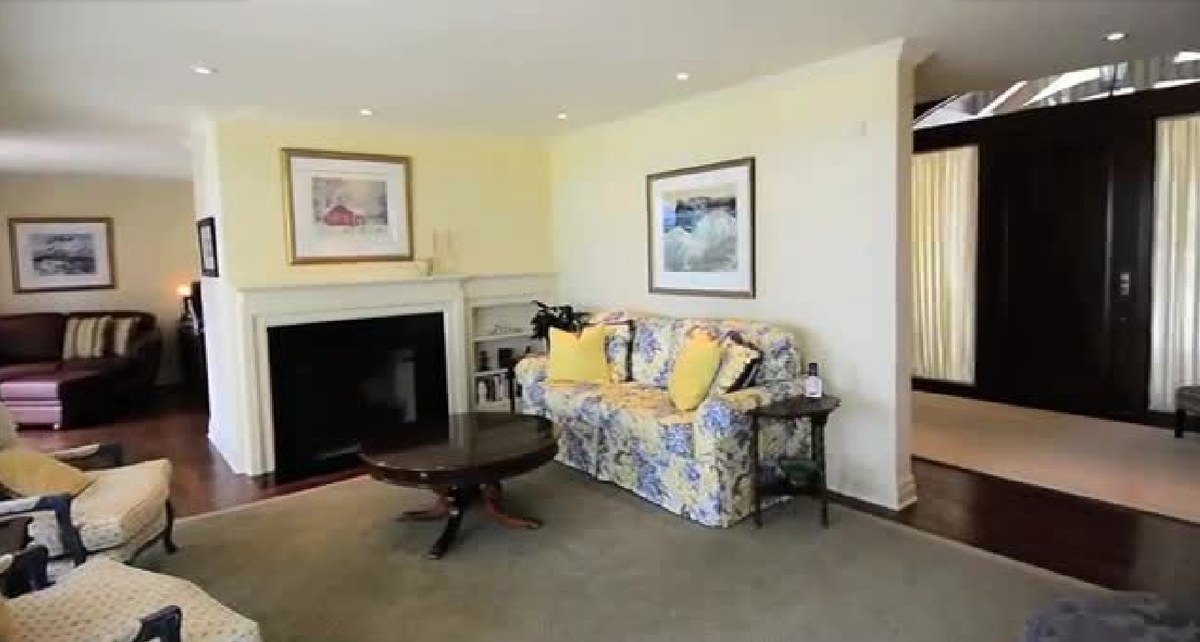}
  \caption{The groundtruth frame corresponding to the query "fireplace".}
  \label{fig:appx_over_cap_gt}
\end{figure}

%% file: fig_inputs/appx_figs_input/fig_reasoning_examples_1.tex
\begin{figure}[t]
  \centering
  \includegraphics[width=\linewidth]{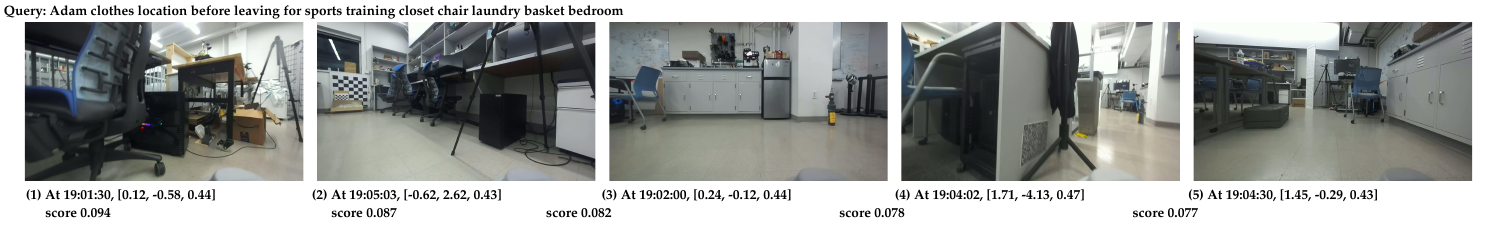}
  \caption{The retrieval tool responds with the images alongside their similarities, positions, and timestamps. The query is \emph{Adam clothes location before leaving for sports training closet chair laundry
basket bedroom}. The experiments are conducted in a lab space.}
  \label{fig:appx_reasoning_examples_1}
\end{figure}

%% file: fig_inputs/appx_figs_input/fig_reasoning_examples_2.tex
\begin{figure}[t]
  \centering
  \includegraphics[width=\linewidth]{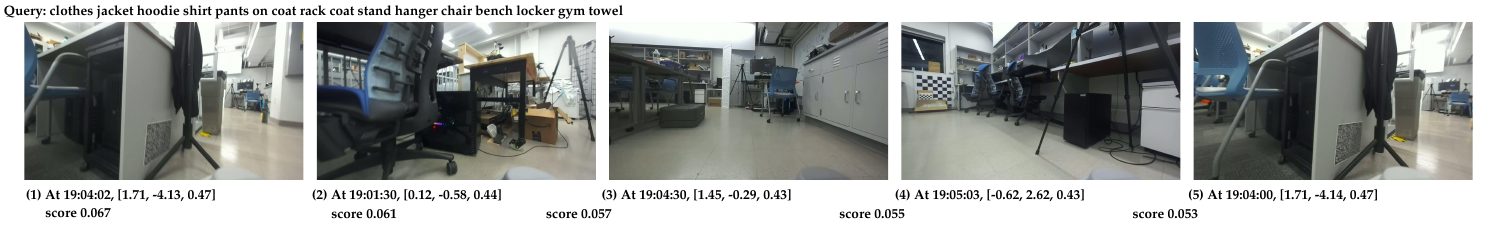}
  \caption{The text retrieval tool's response. The query is \emph{clothes jacket hoodie shirt pants on coat rack coat stand hanger chair bench
locker gym towel}.}
  \label{fig:appx_reasoning_examples_2}
\end{figure}

%% file: fig_inputs/appx_figs_input/fig_reasoning_examples_3.tex
\begin{figure}[t]
  \centering
  \includegraphics[width=\linewidth]{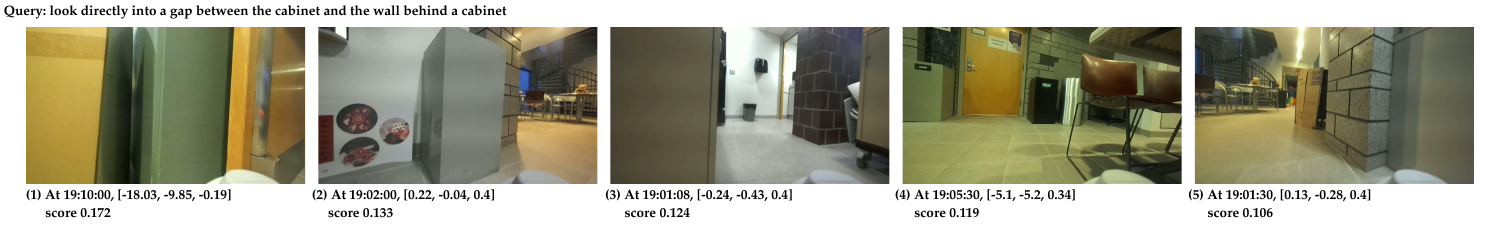}
  \caption{The text retrieval tool's response, experimented in a public space. The query is \emph{look directly into a gap between the cabinet and the wall behind a cabinet}.}
  \label{fig:appx_reasoning_examples_3}
\end{figure}

%% file: fig_inputs/fig_real_world.tex
\begin{figure*}[htbp!]
 \centering
 \includegraphics[width=\textwidth]{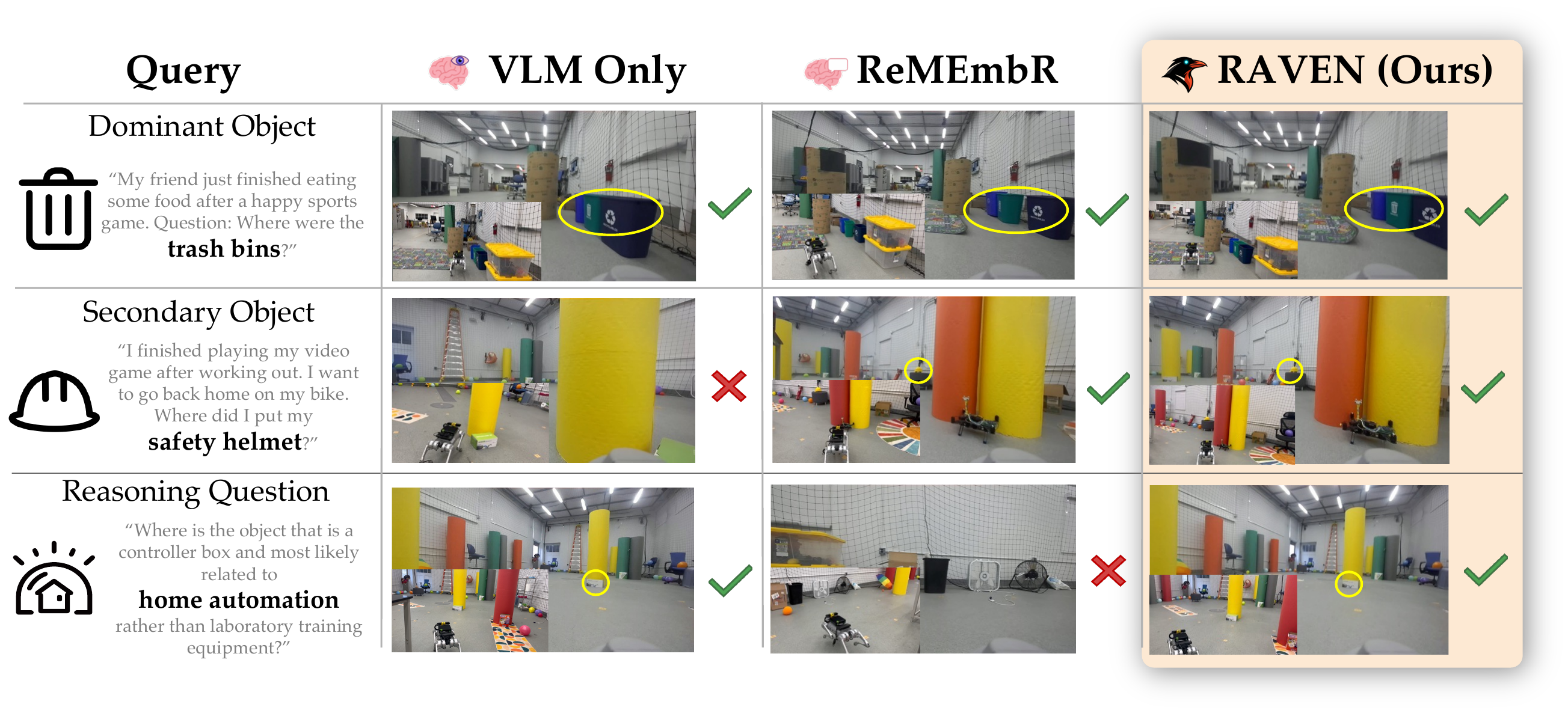}
 \caption{Comparison of three methods, \method{}, \remembr{}, and \vlmonly baseline, across three query types in \ourdataset{}. We show third-person images of the robot together with its ego-view observations corresponding to predicted $(x, y)$ goal coordinates for each method. While the three methods perform comparably on dominant-object queries, \remembr{} fails on the reasoning query and the \vlmonly baseline degrade on secondary-object queries, where \method{} remains reliable. We use Gemini-2.5-Flash as the VLM agent across three methods. See Appendix~\ref{appendix:real_fail_case_study_experiment} for a detailed analysis of failure cases for \remembr{} and the \vlmonly baseline.}
 \label{fig:real_world}
\end{figure*}